\pdfoutput=1
\documentclass{article} %
\PassOptionsToPackage{usenames,dvipsnames,svgnames,table}{xcolor}
\usepackage[preprint]{iclr2027_conference}
\usepackage{iftex}
\ifPDFTeX
  \usepackage{times}
\else
  \usepackage{fontspec}
\fi

\usepackage{graphicx}
\graphicspath{{figs/}}
\usepackage{afterpage}
\usepackage{xspace}
\usepackage{amsmath}
\usepackage{amssymb}
\usepackage{booktabs}
\usepackage{tabularx}
\usepackage{longtable}
\usepackage{multirow}
\usepackage{enumitem}
\usepackage{tikz}
\usetikzlibrary{positioning, arrows.meta, fit, calc}

\usepackage{amsmath,amsfonts,bm}

\def\figref#1{figure~\ref{#1}}

\def\secref#1{section~\ref{#1}}

\def\eqref#1{equation~\ref{#1}}

\def\1{\bm{1}}

\DeclareMathAlphabet{\mathsfit}{\encodingdefault}{\sfdefault}{m}{sl}
\SetMathAlphabet{\mathsfit}{bold}{\encodingdefault}{\sfdefault}{bx}{n}

\usepackage{hyperref}
\usepackage{url}
\usepackage{xcolor} %
\hypersetup{
  colorlinks=true,
  citecolor={MidnightBlue},
  linkcolor={BrickRed},
  urlcolor={MidnightBlue}
}
\usepackage{rotating}
\usepackage{framed}
\usepackage{fvextra}
\definecolor{shadecolor}{gray}{0.94}

\definecolor{d2lcolor}{HTML}{6C3FA0}%
\definecolor{paperA}{HTML}{2A78D6}%
\definecolor{paperB}{HTML}{EB6834}%
\newcommand{\doctolora}{\textcolor{d2lcolor}{D2L}\xspace} %
\def\matrix#1{{\mathbf{#1}}}
\renewcommand{\secref}[1]{Section~\ref{sec:#1}}
\renewcommand{\figref}[1]{Fig.~\ref{fig:#1}}
\newcommand{\tabref}[1]{Table~\ref{tab:#1}}

\usepackage{titlesec}
\titleformat{\subsection}[runin]{\normalfont\bfseries}{\thesubsection}{0.5em}{}[.]
\titlespacing*{\subsection}{0pt}{1.4ex plus .2ex minus .2ex}{0.8em}

\newcommand{\std}[1]{{\scriptsize$\pm$#1}} %

\ificlrpreprint
\title{Doc2LoRA Provides Decodable\\ Representations of Scientific Ideas}
\else
\title{Doc2LoRA Provides Decodable Representations of Scientific Ideas}
\fi

\author{%
Chand Sahil Mansuri\thanks{Equal contribution.}\quad Joel Zachariah\footnotemark[1]\quad Sadamori Kojaku \\
School of Systems Science and Industrial Engineering \\
Binghamton University, Binghamton, NY, USA \\
\texttt{\{cmansur1,jzachariah,skojaku\}@binghamton.edu}
}

\begin{document}

\maketitle

\begin{abstract}
Representing scientific papers as points in a space lets us search for similar papers and inquire about how fields relate to one another and drive innovation.
Beyond search, the vector space of papers invites generation: mixing papers through simple vector operations creates new points, mirroring combinatorial novelty, the recombination of existing ideas into new ones.
However, a mixed point often represents an idea no paper has yet realized, with no papers nearby to identify the idea.
We propose representing each paper by a LoRA adapter generated by the Doc-to-LoRA hypernetwork.
Every point in the space, including mixtures, thus represents a large language model (LLM) open to questions and instructions in natural language.
On papers from the American Physical Society (APS), we instruct the LLM at the average of each subfield to name the field in a few words and obtain labels closer to the official names than the labels of five baselines, as judged by word overlap and a panel of five LLM judges.
We also ask the LLMs at points between two APS papers to write an abstract and obtain descriptions shifting from one paper to the other in step with the mixing weight.
While Doc-to-LoRA is trained for generation, a small invertible transform makes the embeddings competitive for search, on par with SPECTER2 and EmbeddingGemma and close to SBERT.
Because the transform is invertible, every point in the transformed space still maps back to an LLM.
The embeddings thus serve both search and generation, enabling researchers to question the idea at any point in the space as a starting point for generating new ideas.
\end{abstract}

\section{Introduction}
\label{sec:introduction}

Scientific papers form the foundational record of scientific knowledge and are collected in massive digital libraries~\citep{lo2020s2orc,priem2022openalex}.
Understanding the landscape of science across these libraries is essential for identifying the structural patterns that drive innovation and inform science policy.
Embedding models transform each paper into a vector and serve as the central tool for mapping this landscape, enabling literature recommendation, large-scale scientific analysis~\citep{reimers2019sbert,cohan2020specter,su2023instructor,beltagy2019scibert,ostendorff2022scincl,fortunato2018science}, and the prediction of future discoveries~\citep{tshitoyan2019mat2vec,peng2021periodical2vec,krenn2020predicting,sourati2023accelerating}.
However, these embeddings serve search rather than generation and are rarely interpretable on their own.
Because most embedding models are trained to preserve similarity structures for search, they compress text by discarding internal semantics, obscuring the actual ideas represented by individual papers~\citep{weller2025limitations,kuratov2025cramming}.

Since the original text is not recoverable from standard vectors, interpreting the idea of a given point---especially an arithmetic point not paired with an actual paper---relies on heuristics.
Labeling nearest documents~\citep{grootendorst2023keyllm,pham2024topicgpt,wang2023goalex} only summarizes surrounding papers instead of describing the point itself.
Vector-to-text inversion~\citep{morris2023vec2text,song2020leakage,li2023geia} is trained to reconstruct verbatim source text, offering no semantic guarantees for synthetic points constructed in latent space.
Both strategies therefore only approximate the underlying idea.
Alternative approaches build interpretability into vectors by mapping coordinates to sparse concepts~\citep{koh2020concept,bhalla2024splice}, fixed answers~\citep{benara2024qaemb}, or sparse autoencoder features~\citep{oneill2024sae}, but they restrict interpretation to predefined vocabularies.

\afterpage{%
\begin{figure}[t]
\centering
\includegraphics[width=\linewidth]{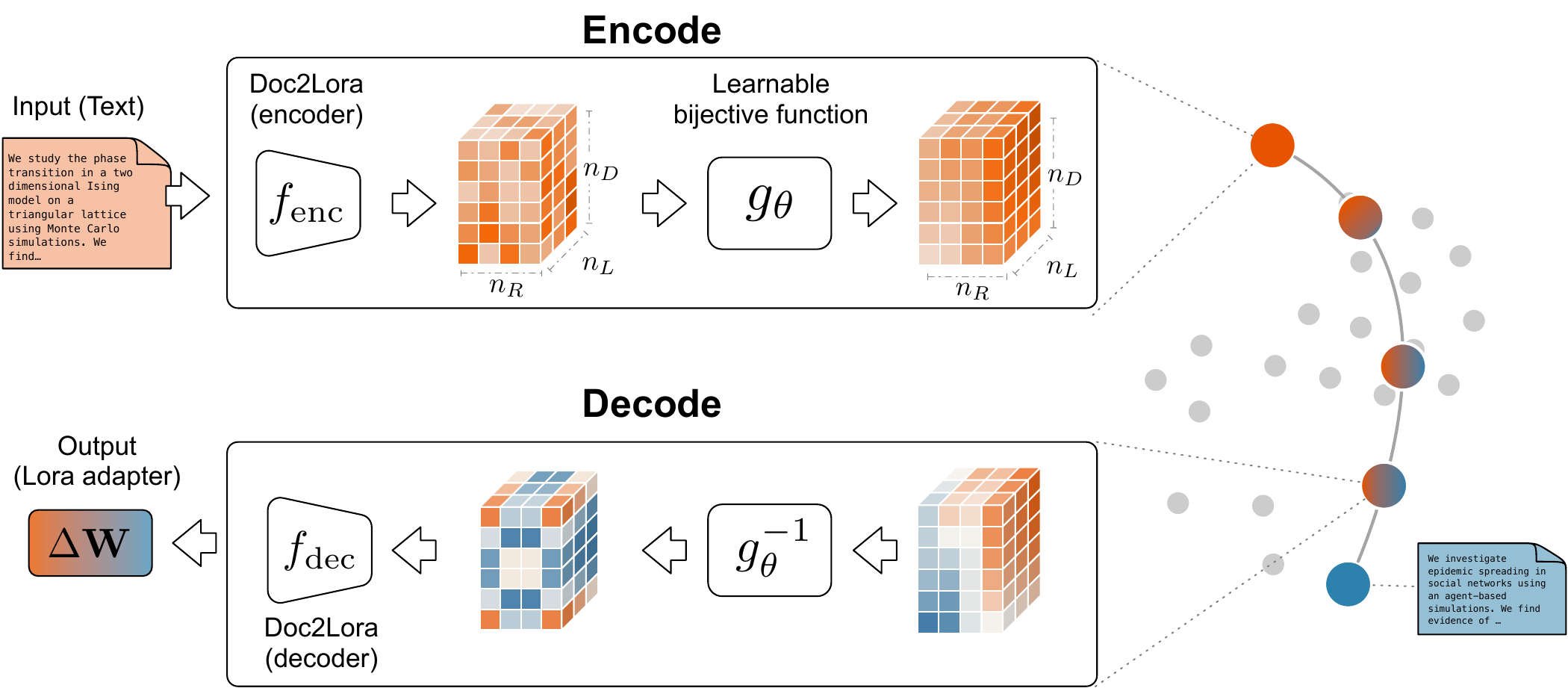}
\caption{
\textbf{Doc2LoRA pipeline.}
\textbf{Encode:} The frozen \doctolora hypernetwork $f_{\mathrm{enc}}$ compresses a document into an embedding tensor $\mathcal{V}$ of shape $(n_L, n_R, n_d)$, sitting between the MLP and LoRA heads (\secref{doc2lora-embedding}). A bijective map $g_\theta$ (\secref{kron-method}) aligns this space for downstream search.
\textbf{Decode:} Any point, including interpolations and averages, is mapped back via $g_\theta^{-1}$ and expanded by $f_{\mathrm{dec}}$ into a LoRA update $\Delta \matrix{W}$, enabling natural language queries to the base model. 
}
\label{fig:method}
\end{figure}%
}

Heuristic labeling and vector decoders offer only passive descriptions of what a point is about.
Generating new ideas requires more than description.
New ideas often arise from recombining existing ideas, a pattern known as combinatorial novelty~\citep{swanson1986undiscovered,uzzi2013atypical,foster2015tradition,youn2015invention,shibayama2021novelty}.
Exploring such combinations calls for active interrogation, asking how two ideas can be combined or what specific mechanism carries over from one idea to another.
Such interrogation requires embeddings that we can query about ideas through dialogue.

We propose conversational embeddings, where every point in the embedding space represents a large language model (LLM) open to questions in natural language.
Because every point in this space is an LLM, performing arithmetic operations---such as averaging a cluster of papers or interpolating between two ideas---yields another LLM that can be queried about the resulting concept. 

We enable this paradigm by leveraging Doc-to-LoRA~\citep{charakorn2026doc2lora}, which extracts a low-dimensional latent representation for each document.
While LLMs are typically parameterized by billions of weights, \doctolora generates a compact adapter that captures only the document-specific changes while shared prior knowledge remains in the frozen base model and is not encoded in the embedding.
We use the compact latent tensor produced by its encoder-decoder architecture as our document embedding.
This representation isolates what makes that document unique, omitting redundant commonalities present across all documents.
Although \doctolora was originally designed for context distillation~\citep{snell2022context}, we show that its latent space has an intuitive geometric structure and is effective as an embedding.

While several works represent texts such as task instructions and documents as LoRA adapters~\citep{chen2025ga,su2025parametricrag,ye2021hypter,mahabadi2021hyperformer,phang2023hypertuning,charakorn2025texttolora}, we select \doctolora~\citep{charakorn2026doc2lora} as our base hypernetwork because it provides both identifiability and scalability.
The same document always maps to the same adapter, and large corpora can be embedded efficiently in a single forward pass.
For example, Parametric RAG~\citep{su2025parametricrag}, which fine-tunes a LoRA adapter per document, lacks identifiability due to random initialization, resulting in different adapters for the same input~\citep{garipov2018loss,frankle2020linear,entezari2022permutation,ainsworth2023gitrebasin}.
GenerativeAdapter~\citep{chen2025ga} achieves identifiability and fast, fixed-size adapter generation, but produces prohibitively large adapters, e.g., $268$M parameters ($537$\,MB in \texttt{bf16}) per document for Mistral-7B---making corpus-level indexing impractical.
In contrast, \doctolora creates a compact, consistent adapter per document in one pass, compressing each to a $147$k-dimensional latent tensor (and poolable to $18$k for search; see \secref{doc2lora-embedding}), achieving both identifiability and scalability.

We test \doctolora embeddings on both generation and search.
On generation, the LLM at the average of a cluster of papers names the field shared by the papers (\secref{cluster-labeling}).
The LLMs at points between two papers describe work combining both (\secref{fusion}).
On search, raw \doctolora embeddings fall behind dedicated text encoders.
A small invertible transform trained on citation pairs brings \doctolora to the level of widely used text encoders while every point still represents an LLM (\secref{similarity}).

In summary, we turn the embedding space of scientific papers from a space for search into a space for both search and generation.
Built on the public SakanaAI \doctolora checkpoint without retraining, conversational embeddings let researchers question any point in the space, including averages and mixtures of papers, in natural language.
A linear map from each embedding to its adapter keeps averaging and mixing well defined (\secref{doc2lora-embedding}).
A small invertible transform aligns the space for search while every point still represents an LLM (\secref{kron-method}).
Together, conversational embeddings let researchers move from finding existing papers to asking about ideas no paper has yet realized, a step toward exploring combinatorial novelty across the scientific literature.
\ificlrpreprint
The code and the workflow that reproduces every result are available at \url{https://github.com/skojaku/doc2lora-embedding}.
\fi

\section{Methods}
\label{sec:method}

\subsection{Doc-to-LoRA hypernetwork}
\label{sec:doc2lora}

\doctolora is a hypernetwork that converts a document $x$ to an LoRA adapter in one forward pass.
Specifically, a document $x$ is first sent to the base LLM, which produces a sequence of hidden states for all tokens in the document.  
These hidden states are then sent to a perceiver transformer~\citep{jaegle2021perceiver}, which compresses into a fixed-size tensor $\mathcal{U} \in \mathbb{R}^{n_L \times n_R \times n_d}$, where $n_L$ is the number of layers in the base LLM, $n_R$ is the number of rank slots, and $n_d$ is the dimension of the hidden states.
This fixed-size tensor $\mathcal{U}$ is then sent to an MLP followed by row-wise $L_2$ normalization, generating another fixed-size tensor $\mathcal{V} \in \mathbb{R}^{n_L \times n_R \times n_d}$.
Finally, a linear head $f_{\mathrm{dec},\ell}$ maps $\mathcal{V}_{\ell}$ to the LoRA adapter matrices, independently for each layer and with no weight sharing.

The resulting LoRA adapter consists of a pair of matrices $\matrix{A}_{\ell} \in \mathbb{R}^{m_\ell \times n_R}$ and $\matrix{B}_{\ell} \in \mathbb{R}^{n_R \times k_\ell}$ for each layer $\ell$, where $n_R$ is the rank of the LoRA adapter. 
These matrices $\matrix{A}_{\ell}$ and $\matrix{B}_{\ell}$ are then added to the linear layer $\matrix{W}_{\ell}$ of the MLP in the base LLM during inference.
Loading the resulting updates $\{\Delta\matrix{W}_{\ell}\}$ into the frozen base model yields a document-conditioned model that answers as if $x$ were in its context. We use the public SakanaAI checkpoint on Qwen3-4B~\citep{qwen3-2025} in the main text and report other base models in Appendix~\ref{app:benchmarks}.
The faithfulness of the decoded output depends on the hypernetwork, as evaluated by \citet{charakorn2026doc2lora}.
We take this as given and focus on the embedding space itself. 
Paraphrasing the prompt changes the wording but rarely the content of the responses (Appendix~\ref{app:prompt-sensitivity}).

\subsection{\doctolora embedding}
\label{sec:doc2lora-embedding}

We leverage the post-MLP tensor $\mathcal{V}$ as the document embedding, as the following linear head $f_{\mathrm{dec},\ell}$ maps it directly to the LoRA adapter matrices, preserving linear equivariance:
\begin{equation}
f_{\mathrm{dec},\ell}\Big(\sum_i w_i\, \matrix{V}_{i,\ell}\Big) = \sum_i w_i\, f_{\mathrm{dec},\ell}(\matrix{V}_{i,\ell}).
\label{eq:linearity}
\end{equation}
Averaging or interpolating embeddings is therefore equivalent to performing the same operation on the factor matrices $\matrix{A}_{\ell}$ and $\matrix{B}_{\ell}$.
This equivalence does not apply to the adapter updates $\Delta\matrix{W}_{\ell} = \matrix{A}_{\ell}\matrix{B}_{\ell}$ because the product is not linear in the embedding.
This linearity is lost if the embedding is taken from the nonlinear pre-MLP matrices.
Each $\mathcal{V}$ is a third-order tensor, with $\ell$-th slice $\matrix{V}_{\ell} \in \mathbb{R}^{n_R \times n_d}$ containing $L_2$-normalized, $n_d{=}512$-dimensional unit vectors per $n_R{=}8$ rank (\figref{method}).

When restricted to search, we can downsize the hidden tensor $\mathcal{V}$ while preserving much similarity information by taking the mean over the rank dimension $n_R$ (Spearman $\rho$ of $.97$--$.98$ between the full-tensor and pooled cosine similarities for Qwen3-4B, on $500$ documents from each evaluation corpus (Appendix~\ref{app:data})).
In our paper, we use the mean-pooled tensor as embedding for similarity search discussed in \secref{similarity}, while use the full tensor for generation tasks. Unless otherwise noted, we use the full tensor.

\subsection{Decoding documents and interpolants}
\label{sec:decode}

We decode an embedding by expanding $\mathcal{V}$ into a LoRA adapter and querying the frozen base model with no document in the prompt.
We illustrate with two short recipes, an Italian cacio e pepe and a Japanese kake udon, encoded with the Mistral-7B hypernetwork and decoded in full-rank mode (Appendix~\ref{app:recipe-fusion}).
With the Italian adapter loaded, the query ``What cheese is used?'' returns ``The cheese used in this recipe is Pecorino Romano.''
Next, we interpolate the two embeddings as $\mathcal{V}(\alpha)=(1-\alpha)\mathcal{V}_A+\alpha\mathcal{V}_B$, where $\mathcal{V}_A$ and $\mathcal{V}_B$ are the embeddings of the two recipes.
The endpoints $\alpha=0$ and $\alpha=1$ decode to cacio e pepe and kake udon, each with minor paraphrases.
At $\alpha=0.5$ the decode is ``Kake Udon with Creamy Parmesan Sauce.''
The dish coats udon noodles in a melted-cheese sauce and seasons them with mirin and soy.
Parmesan replaces Pecorino Romano.
The decode also adds heavy cream, which is absent from both sources.
We decode greedily, and the examples we report are single runs. 

\subsection{Adapting the geometry with a bijective transform}
\label{sec:kron-method}

The public hypernetwork is trained for generation.
We refine its tensor for search without retraining the base LLM and hypernetwork by adding a linear layer $g_\theta$ on top of the embedding, with a bijectivity constraint.
More specifically, for each layer $\ell$, we transform $\matrix{V}_{\ell}\in\mathbb{R}^{n_R\times n_d}$ by
\begin{align}
   \label{eq:transform}
   \matrix{V}'_{\ell} = g_\theta(\matrix{V}_{\ell}) = a_\ell \matrix{V}_{\ell} \matrix{C},
\end{align}
where $a_\ell>0$ controls the scaling of layer $\ell$, and $\matrix{C}\in\mathbb{R}^{n_d\times n_d}$ is an invertible linear matrix.
To ensure invertibility, we factor $\matrix{C}$ as $\matrix{R} \text{diag}(\sigma)$, where $\matrix{R}\in\mathbb{R}^{n_d\times n_d}$ is a rotation matrix and $\text{diag}(\sigma)\in\mathbb{R}^{n_d\times n_d}$ is a diagonal matrix with positive entries $\sigma_i>0$ (Appendix~\ref{app:bijective}).
This parametrization uses only $n_L + n_d^2$ parameters ($262$k for Qwen) and brings the embedding on par with dedicated text encoders on the similarity benchmarks we tested.
We train $g_\theta$ with InfoNCE~\citep{oord2018infonce} on citation pairs (Appendix~\ref{app:bijective}) using the OpenAlex citation pairs (negatives in-batch, as in \texttt{SPECTER2}), with the base model frozen.

\section{Results}
\label{sec:results}

\subsection{Averaging embeddings names the field a cluster shares}
\label{sec:cluster-labeling}

\begin{figure}[t]
\centering
\includegraphics[width=\linewidth]{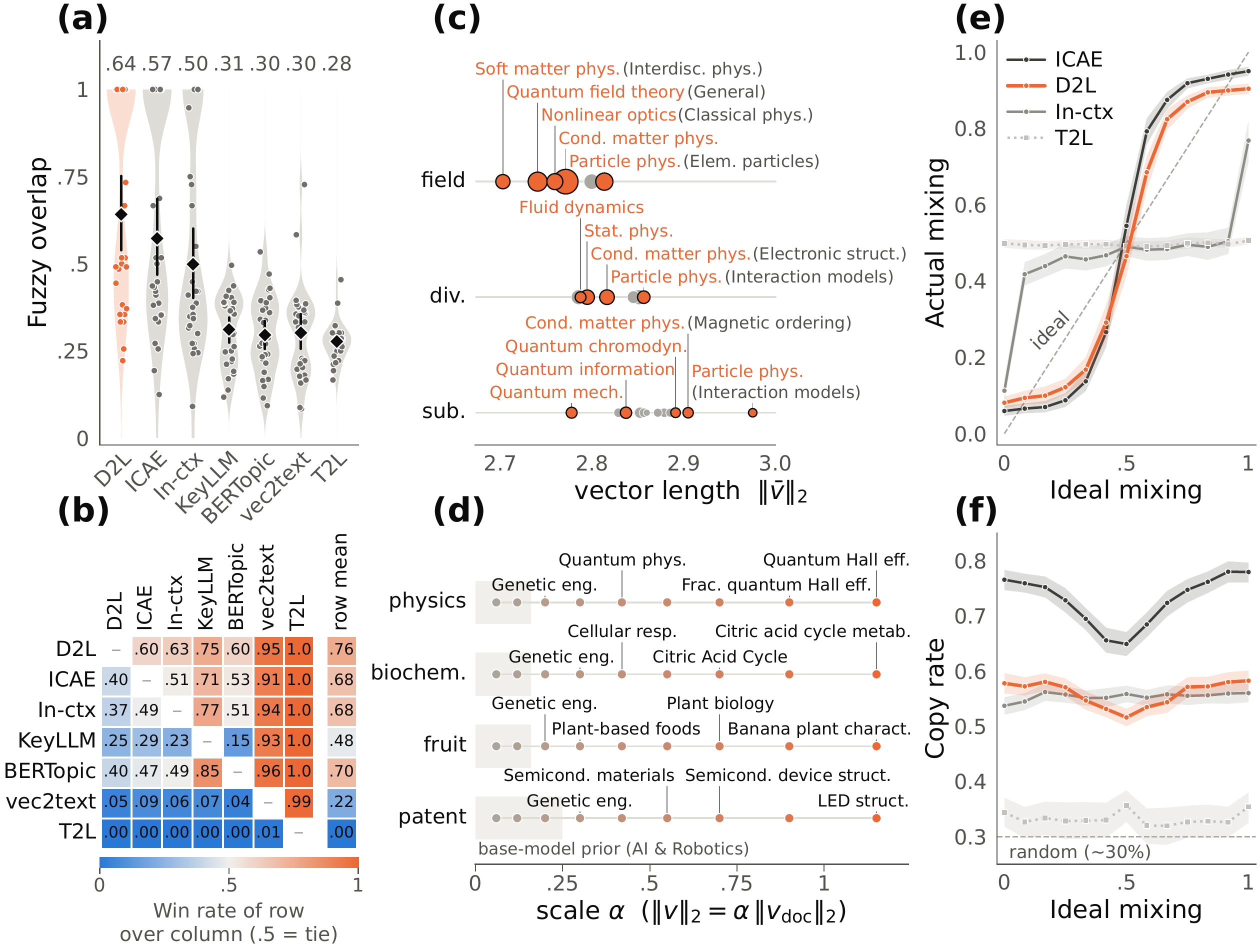}
\caption{
    Cluster labeling and mixing with \doctolora.
    \textbf{(a)} Fuzzy token-set overlap between each decoded label and official 28 PACS labels. Each point is a node, with distribution represented by kernel density estimation (KDE).
    \textbf{(b)} Pairwise judge win rates, showing the fraction of PACS labels where the generated label from the row method is judged closer to the official label than that from the column. Color runs blue (row loses), grey (tie), to orange (row wins), row means at right.
    \textbf{(c)} Cluster centroids, decoded per PACS field/division/subdivision. Horizontal axis is vector length, circle area is the number of papers. Decoded labels (orange) and official PACS names (grey, if different) are shown.
    \textbf{(d)} Specificity as a function of vector length: longer vectors yield specific, shorter more general concepts. For each Wikipedia seed, lines show continuity of label, grey bands mark collapse to base-model prior.
    \textbf{(e)} Decoded abstract position (SBERT space) as a function of B weight, A$\to$B, for 50 near pairs (papers from the same PACS subtopic). \doctolora and \texttt{ICAE} track the target, in-context stays near midpoint, \texttt{T2L} flat.
    \textbf{(f)} Verbatim copy rate by method, measured as the fraction of output tokens present in sources. \texttt{ICAE} is highest, and \doctolora and in-context prompting are comparable.
}
\label{fig:cluster-labels}
\end{figure}

A widespread practice for intepretating clustering results is either manual interpretation of the members, or running a keyword or LLM pipeline that needs careful tuning before it returns a usable name~\citep{grootendorst2020keybert,grootendorst2022bertopic,grootendorst2023keyllm,murray2025labeling}.
With \doctolora, labeling becomes as simple as averaging the vectors of a cluster and decoding the mean (\figref{cluster-labels}c).

We validate cluster labeling on physics papers from the American Physical Society (APS), using the Physics and Astronomy Classification Scheme (PACS) (Appendix~\ref{app:data}).
PACS is a hierarchical classification system for topics in physics, in which each topic is assigned a numeric code. 
For example, ``03.65'' is assigned to quantum mechanics.
If the last two digits are omitted, it refers to the broader, higher-level topic above it. For example, ``03'' refers to quantum mechanics and quantum field theory, while ``0'' refers to general physics.
Authors assign codes to their papers to organize them, with the first code indicating the paper's main topic.
We group papers based on their first code at each of the three levels of the hierarchy (field, division, subdivision) and calculate the average of the vectors within each group.
We decode these averages using the instruction ``In 2 to 3 words, name the scientific field that all of these documents belong to'' and compare the decoded labels with the official PACS names of the nodes (Appendix~\ref{app:baseline-labels}).

We compare \doctolora to five baselines: \texttt{KeyLLM}~\citep{grootendorst2023keyllm}, which prompts an LLM with the original documents to extract keywords; \texttt{BERTopic}~\citep{grootendorst2022bertopic}, which represents each given cluster by its most distinctive words (c-TF-IDF); \texttt{vec2text}~\citep{morris2023vec2text}, which iteratively edits text to match a target embedding; \texttt{ICAE}~\citep{ge2024icae}, which encodes documents into memory slots and decodes them back; and \texttt{T2L}~\citep{charakorn2025texttolora}, which generates LoRA adapters from a frozen \texttt{GTE} encoder~\citep{li2023gte}. 
We also compare against the base model's own hidden states (see Appendix~\ref{app:actpatch} for details). 
All methods receive the same input: unmodified paper titles and abstracts.

We quantify the quality of the decoded labels by fuzzy token-set overlap, a score from $0$ to $1$ with no LLM in the loop (\figref{cluster-labels}a).
\doctolora leads with $.642 \pm .055$, ahead of \texttt{KeyLLM}, \texttt{BERTopic}, \texttt{vec2text}, \texttt{ICAE}, and \texttt{T2L}.
As an additional validation, we use five LLM judges (\texttt{glm-5.3-flash}, \texttt{gemini-3.8-flash}, \texttt{gpt-5.6-luna}, \texttt{muse-spark-1.3}, and \texttt{mistral-medium-3.1}) to compare every pair of methods on each node (\figref{cluster-labels}b).
LLM judges consistently prefer \doctolora's labels over all baselines.
A control rules out that any cluster draws a plausible field name: on clusters sampled at random across PACS chapters the judges accept no label at all, against $73.3\%$ for the real nodes (Appendix~\ref{app:incoherent}).

Upon manual inspection, \doctolora decodes cluster means into formal field names such as ``particle physics'' and ``condensed matter physics'' more frequently than all baselines (\figref{cluster-labels}c).
The baselines exhibit four main failure modes.
\texttt{T2L} produces a handful of unrelated, fixed field names (e.g., ``Social psychology'') for all nodes; \texttt{vec2text} generates indecipherable pseudo-words (e.g., ``inferromagnetic spin-diabetic scattering interactions''). \texttt{KeyLLM} returns overly specific keyword lists from few medoid papers, and \texttt{BERTopic} returns lists of topical words (``quark, mass, model, gauge, higgs, $\ldots$'') but never the name of the field. \texttt{ICAE} decodes plausible field names labels are too broad (``Materials science'') or too specific (``Quantum many-body systems''). 
The differing performance of \doctolora and \texttt{T2L} shows that just creating a LoRA adapter is not enough. 
Decodable clustering depends on how the adapter is trained.

\doctolora avoids generating cluster names that are too broad or too narrow because it interprets the length of the mean vector as an indicator of abstraction level. 
When document vectors point in the same direction (i.e., are well aligned), their average keeps much of the original length. 
If they vary widely, averaging pulls the mean closer to zero, shortening the vector. For \doctolora, we observe that the average vector length grows as we move from general categories (top-level fields) to more detailed ones (fine-grained subdivisions; \figref{cluster-labels}c).
While the contraction of cluster means is a general geometric property of averaging diverse vectors, only \doctolora translates vector length into the appropriate level of abstraction when decoding names.
By contrast, sweeping vector length in \texttt{ICAE} leaves decoded labels largely unchanged, and \texttt{vec2text} remains specific until collapsing at low scaling without producing broader field names (Appendix~\ref{app:length-dial}).
To test this mechanism, we embed and deliberately rescale Wikipedia introductions (\figref{cluster-labels}d). 
With high scaling ($\alpha$), we get detailed concepts such as ``Citric acid cycle metabolism.'' 
As we reduce $\alpha$, the decoded label shifts to something more general, like ``Cellular respiration.'' 
For Nikola Tesla, shrinking the vector gives ``AC power.'' For a light-emitting diode, it shifts to ``Semiconductor device structure'' (Appendix~\ref{app:length-dial}). 
Extremely short vectors produce generic names from the base model, as if the LoRA adapter were absent. 
If the vector is too long, the output fails to decode because it drifts off the space of meaningful representations.

\subsection{Midpoints decode to blends of both sources}
\label{sec:fusion}

\begin{table}[t]
    \centering
    \footnotesize
    \setlength{\tabcolsep}{4pt}
    \caption{One near pair (two papers from the same PACS subtopic) decoded along the interpolation path at $3/12$, $5/12$, and $9/12$ weight on paper B, under the prompt quoted in the text. $5/12$ is the grid point where the \doctolora decode uses the terms of both papers.
    Paper A is ``Statistical mechanics of learning from examples'' (PhysRevA 45.6056) and paper B is ``Good quantum error-correcting codes exist'' (PhysRevA 54.1098).
    Each cell is the opening of the decoded abstract, cut at $42$ words. Terms from paper A are in \textcolor{paperA}{blue} and terms from paper B in \textcolor{paperB}{orange}.
    }
    \label{tab:mixing-decode}
    \providecolor{paperA}{HTML}{2A78D6}
\providecolor{paperB}{HTML}{EB6834}
\resizebox{\linewidth}{!}{%
\begin{tabular}{@{}l *{3}{>{\raggedright\arraybackslash}p{0.36\linewidth}}@{}}
\toprule
& \multicolumn{3}{c}{Ideal mixing weight on paper B} \\
\cmidrule(lr){2-4}
Method & $3/12$ & $5/12$ & $9/12$ \\
\midrule
\doctolora & This research investigates the learning of real-valued functions in \textcolor{paperA}{neural networks} using a \textcolor{paperA}{statistical-mechanical} framework, focusing on the \textcolor{paperA}{generalization error} of trained networks. The problem involves understanding how well a network can generalize from a finite \textcolor{paperA}{training set} to unseen data, particularly\ldots{} & This research investigates the learning of quantum gates in a unitary model of \textcolor{paperB}{quantum computation}, where a quantum system evolves under a unitary transformation and is trained using a quantum \textcolor{paperA}{neural network}\ldots{} & This research topic investigates the existence and properties of \textcolor{paperB}{quantum error-correcting codes}, which are essential for protecting \textcolor{paperB}{quantum information} from \textcolor{paperB}{decoherence}. The problem is to determine whether, given a quantum system of \textcolor{paperB}{qubits}, there exists a unitary encoding (quantum map) that transforms\ldots{} \\
\texttt{ICAE} & This study investigates the learning of \textcolor{paperA}{statistical mechanics} from \textcolor{paperA}{feedforward neural networks}, focusing on the existence of stochastic, \textcolor{paperA}{Gibbs}-free energy-based models. The study assumes a temperature-dependent distribution of networks and considers both realizable and unrealizable rules\ldots{} & This research explores the existence of \textcolor{paperB}{quantum error-correcting codes} in \textcolor{paperA}{statistical mechanics}, specifically in the context of learning from examples. The study focuses on the existence of a unitary random code, which is a type of \textcolor{paperB}{error-correcting code} that can be defined\ldots{} & This research explores the existence of \textcolor{paperB}{quantum error-correcting codes} (\textcolor{paperB}{QECCs}) in the context of \textcolor{paperB}{quantum computing}. It defines a unitary mapping called a \textcolor{paperB}{quantum error-correcting code} (\textcolor{paperB}{QECC}) that maps \textcolor{paperB}{qubits} to a subspace of the quantum state space\ldots{} \\
in-context & This research investigates the \textcolor{paperA}{statistical-mechanical} behavior of quantum \textcolor{paperA}{neural networks} under the influence of both stochastic learning dynamics and quantum \textcolor{paperB}{decoherence}, blending the \textcolor{paperA}{statistical mechanics of learning from examples} (A:75\%) with the principles of \textcolor{paperB}{quantum error correction} (B:25\%)\ldots{} & This research investigates the \textcolor{paperA}{statistical-mechanical} behavior of quantum \textcolor{paperA}{neural networks} under the influence of both stochastic training and quantum \textcolor{paperB}{decoherence}, blending the framework of learning from examples (A) with the principles of \textcolor{paperB}{quantum error correction} (B) in a 58:42 proportion\ldots{} & This research investigates the emergence of quantum \textcolor{paperA}{spin-glass} phases in \textcolor{paperA}{neural network} architectures under the influence of quantum \textcolor{paperB}{decoherence}, blending \textcolor{paperA}{statistical mechanics of learning} (A) with the theory of \textcolor{paperB}{quantum error correction} (B)\ldots{} \\
\bottomrule
\end{tabular}}

\end{table}

\doctolora can decode text from any point in its embedding space, including points between documents. 
We test this by interpolating between APS paper pairs with controlled PACS taxonomy distance: 50 ``near'' pairs from the same subtopic, and 50 ``far'' pairs from different subtopics, both sampled randomly.
For each pair, we decode at 13 points along the interpolation path between documents using \doctolora, in-context prompting, \texttt{ICAE}, and \texttt{T2L}, all under the prompt \textit{``Write a detailed abstract (four to six sentences) describing this research topic: its problem, methods, and findings.''} (\tabref{mixing-decode}; pair sampling, per-method interpolation, and metrics in Appendix~\ref{app:edge-protocol}).
In-context prompting has an advantage, as it directly receives both endpoint documents and mixing proportions, while \doctolora, \texttt{ICAE}, and \texttt{T2L} only receive source content indirectly via parameters, e.g., LoRA weights or memory slots. 

We evaluate similarity to endpoints using SBERT-based scores.
We also calculate the verbatim copy rate~\citep{grusky2018newsroom}, which is the fraction of output tokens shared with either source.
While \doctolora and \texttt{ICAE} follow the true mixing weights along the interpolation path, in-context prompting remains near the midpoint, except at the endpoints (\figref{cluster-labels}e).
The copy rate separates \texttt{ICAE} from both alternatives.
\texttt{ICAE} reuses source wording at $.73$ on either stratum, while \doctolora ($.56$) and in-context prompting ($.52$--$.55$) stay close to each other (\figref{cluster-labels}f).
\texttt{T2L} collapses onto a few input-independent texts.
Its decoded position stays within $.490$--$.506$ across all $13$ points, endpoints included, while \doctolora moves from $.081$ to $.904$ (\figref{cluster-labels}e).
The endpoints are therefore not decodable, and interpolation is moot.

The separation between methods is robust to prompt wording.
Decoding every interpolation point again under three paraphrases of the instruction leaves the tracking of \doctolora, \texttt{ICAE}, and in-context prompting unchanged (Appendix~\ref{app:prompt-sensitivity}).

\subsection{Retrieval with \doctolora embeddings}
\label{sec:similarity}

\doctolora is designed to induce specific behaviors in LLMs, which may or may not contain topic-level information about each document.
If \doctolora adapters do encode topics within their representations, and if we can extract and align this information geometrically, \doctolora has practical value as an embedding for retrieval tasks.
To test the capacity of \doctolora embeddings for retrieval, we apply a simple linear transformation to the \doctolora embedding space, aiming to make topic similarity more directly accessible while preserving decodability, and then evaluate its effectiveness on standard retrieval benchmarks.

We adapt \doctolora embeddings with a lightweight linear transformation $g_\theta$, containing $262$k parameters for Qwen.
We train $g_\theta$ on $42{,}332$ uniformly-sampled OpenAlex citation pairs, representing less than $0.002\%$ of the $2.56$B possible pairs.
This supervision rebalances the representation for similarity tasks without sacrificing adapter decodability.
We compare \doctolora embeddings with and without $g_\theta$ against commonly used text encoders---\texttt{SPECTER2}~\citep{singh2023scirepeval}, \texttt{Instructor}~\citep{su2023instructor}, \texttt{EmbeddingGemma}~\citep{embeddinggemma2025}, \texttt{SBERT}~\citep{reimers2019sbert}, \texttt{GTE}~\citep{li2023gte} on the following tasks (Appendix~\ref{app:data} for data details and Appendix~\ref{app:benchmarks} for the full benchmark protocol).

\begin{table}[t]
\centering
\small
\caption{Similarity benchmarks, absolute scores. Each row is one task on one field or dataset; \emph{raw} is the \doctolora embedding and \emph{adapted} applies $g_\theta$, the linear transform trained on OpenAlex. 
Text encoders are ordered by mean rank. 
Cells are the bootstrap mean $\pm$ its standard deviation over $1{,}000$ resamples. 
Bold and shaded indicates the top method per row; the second-best is underlined. For disambiguation, \texttt{SPECTER} replaces \texttt{SPECTER2}. Full protocol: Appendix~\ref{app:benchmarks}.}
\label{tab:similarity}
\setlength{\tabcolsep}{4pt}
\renewcommand{\arraystretch}{1.05}
\resizebox{\textwidth}{!}{%
\begin{tabular}{@{}llccccccc@{}}
\toprule
 & & \multicolumn{2}{c}{\doctolora} & \multicolumn{5}{c}{text encoders} \\
\cmidrule(lr){3-4}\cmidrule(l){5-9}
Task & Field / dataset & raw & adapted & \texttt{SBERT} & \texttt{GTE} & \texttt{Emb.Gemma} & \texttt{Instructor} & \texttt{SPECTER2} \\
\midrule
\multirow{3}{*}{Next-paper (AUC)}
 & Economics & .813\std{.001} & \cellcolor{gray!10}\underline{.910}\std{.001} & \cellcolor{gray!25}\textbf{.937}\std{.001} & \cellcolor{gray!10}\underline{.910}\std{.001} & .906\std{.001} & .877\std{.001} & .903\std{.001} \\
 & Psychology & .799\std{.001} & \cellcolor{gray!10}\underline{.923}\std{.001} & \cellcolor{gray!25}\textbf{.937}\std{.001} & .912\std{.001} & .910\std{.001} & .888\std{.001} & .908\std{.001} \\
 & Physics & .880\std{.001} & \cellcolor{gray!10}\underline{.958}\std{.001} & .956\std{.001} & \cellcolor{gray!25}\textbf{.962}\std{.001} & .947\std{.001} & .908\std{.001} & .941\std{.001} \\
\midrule
\multirow{3}{*}{Topic (macro-F1)}
 & Economics & .242\std{.015} & .374\std{.022} & \cellcolor{gray!25}\textbf{.418}\std{.022} & \cellcolor{gray!10}\underline{.398}\std{.023} & .362\std{.021} & .379\std{.021} & .344\std{.021} \\
 & Psychology & .217\std{.010} & .340\std{.020} & \cellcolor{gray!25}\textbf{.366}\std{.023} & .350\std{.019} & \cellcolor{gray!10}\underline{.358}\std{.020} & .348\std{.026} & .319\std{.020} \\
 & Physics & \cellcolor{gray!10}\underline{.590}\std{.012} & \cellcolor{gray!10}\underline{.590}\std{.009} & .573\std{.011} & .587\std{.010} & .588\std{.010} & .574\std{.010} & \cellcolor{gray!25}\textbf{.598}\std{.007} \\
\midrule
\multirow{3}{*}{Collaboration (AUC)}
 & Economics & .596\std{.011} & .629\std{.012} & \cellcolor{gray!25}\textbf{.655}\std{.011} & .627\std{.012} & \cellcolor{gray!10}\underline{.633}\std{.012} & .620\std{.012} & .632\std{.011} \\
 & Psychology & .602\std{.012} & \cellcolor{gray!10}\underline{.687}\std{.010} & \cellcolor{gray!25}\textbf{.691}\std{.011} & .668\std{.011} & .666\std{.011} & .649\std{.011} & .665\std{.011} \\
 & Physics & .792\std{.004} & .795\std{.004} & .832\std{.003} & \cellcolor{gray!10}\underline{.839}\std{.003} & .828\std{.003} & \cellcolor{gray!25}\textbf{.854}\std{.003} & .826\std{.004} \\
\midrule
\multirow{5}{*}{Author disamb.\ (B\textsuperscript{3} F1)}
 & zbMATH & .933\std{.001} & .932\std{.001} & \cellcolor{gray!25}\textbf{.944}\std{.001} & \cellcolor{gray!10}\underline{.939}\std{.001} & .934\std{.002} & .936\std{.001} & .934\std{.001} \\
 & QIAN & .787\std{.005} & .850\std{.004} & \cellcolor{gray!10}\underline{.865}\std{.004} & \cellcolor{gray!25}\textbf{.866}\std{.004} & .832\std{.004} & .832\std{.004} & .832\std{.004} \\
 & ArnetMiner & .678\std{.004} & \cellcolor{gray!25}\textbf{.708}\std{.005} & .698\std{.005} & .698\std{.004} & \cellcolor{gray!10}\underline{.706}\std{.004} & .679\std{.004} & .669\std{.004} \\
 & PubMed & .693\std{.007} & .815\std{.006} & \cellcolor{gray!25}\textbf{.832}\std{.006} & .792\std{.006} & \cellcolor{gray!10}\underline{.822}\std{.006} & .801\std{.006} & .739\std{.006} \\
 & KISTI & .708\std{.002} & .767\std{.002} & \cellcolor{gray!25}\textbf{.793}\std{.002} & \cellcolor{gray!10}\underline{.785}\std{.002} & .771\std{.002} & .754\std{.002} & .746\std{.002} \\
\bottomrule
\end{tabular}
}

\end{table}

With $g_\theta$, \doctolora sits inside the band of the text encoders on all four retrieval tasks. It performs better than \texttt{Instructor} and \texttt{SPECTER2}, is comparable to \texttt{EmbeddingGemma} and \texttt{GTE}, and follows \texttt{SBERT}, which leads on 9 of 14 benchmarks. The average ranks are: \texttt{SBERT} $2.0$, \texttt{GTE} $2.9$, \doctolora{+}$g_\theta$ $3.4$, \texttt{EmbeddingGemma} $3.5$, \texttt{Instructor} $4.7$, \texttt{SPECTER2} $5.0$, and raw \doctolora $6.5$. All text encoders use default settings without task-specific tuning. We do not claim \doctolora outperforms current encoders, but it retains similarity information comparable to standard methods.
Training the same $g_\theta$ in each baseline space improves those encoders marginally or degrades them, with the largest average gain over the $14$ benchmarks being $+.009$ and \texttt{SBERT} losing $.007$.
The transform therefore recovers geometry that the \doctolora space holds without exposing it to cosine similarity, and it gives the baselines no sizeable advantage (Appendix~\ref{app:symmetric}).

These findings highlight an asymmetry between generation-trained and similarity-trained representations.
A strong retrieval encoder does not inherently yield a decodable adapter.
\texttt{T2L} reads \texttt{GTE}, yet it fails to decode documents or support idea arithmetic (\secref{cluster-labeling}, \secref{fusion}).
Conversely, generation-trained representations can implicitly encode topic geometry even when it is not directly exposed as cosine similarity.
Applying $g_\theta$ brings this geometry to the surface in \doctolora, achieving competitive retrieval performance without losing decodability, while offering no meaningful gain to dedicated text encoders.

\section{Related work}
\label{sec:related}

The landscape of document representation has evolved from static text encoders~\citep{reimers2019sbert,su2023instructor,cohan2020specter,singh2023scirepeval}, which shape papers into geometric spaces where points can be compared, but meaningful text recovery from those points remains limited. Traditional approaches rely on dedicated decoders---either trained a posteriori, as in \texttt{vec2text}~\citep{morris2023vec2text}, which constrains outputs to short, often incoherent rephrasings for points off the training manifold~\citep{kuratov2025cramming}, or learned jointly via variational autoencoders~\citep{bowman2016generating,li2020optimus,gomez2018automatic}. The latter enable decoding from any point but produce only brief texts and lack the ability to follow instructions, fundamentally constraining interpretability. Our approach with \doctolora differs: by decoding through a frozen LLM, the embedding gains not only fluency but also instruction-following, making every point in the space \emph{conversationally} interpretable.

Parallel research inserts document content directly into generators. With RAG~\citep{lewis2020rag} and compressed prompt engineering~\citep{jiang2023llmlingua,li2023selective,xu2024recomp}, text remains manipulable only as sequences in the prompt and cannot be averaged or meaningfully composed in vector space. Soft compression---through learned input tokens~\citep{lester2021power,li2021prefix,mu2023gist,chevalier2023autocompressor,ge2024icae,li2024fivehundredx,rau2024cocom,cheng2024xrag} or LoRA adapters~\citep{chen2025ga,su2025parametricrag,charakorn2026doc2lora}---enables vector arithmetic, interpolation, and comparison, and the geometry of these learned vectors tracks meaningful task transfer~\citep{vu2022spot}. Another line of research proposes methods to read the hidden states of a model as text without any learned compression. An example method is Patchscopes~\citep{ghandeharioun2024patchscopes}, which overwrites a hidden state at a target position and prompts the model to decode it.
SelfIE~\citep{chen2024selfie} interprets individual token embeddings through the same mechanism, and LatentQA~\citep{pan2024latentqa} trains a small encoder to answer natural-language questions about activations. These hidden-state methods require no adapter and no training on the document. Therefore, we consider them as an important reference for whether a learned LoRA representation adds value over what the model already knows (Appendix~\ref{app:actpatch}).
Building on this, our work uses the generated LoRA adapter as the embedding itself, demonstrating that its structure is smooth, expressive, and admits both semantic retrieval and idea synthesis (\secref{similarity}, \secref{fusion}).

At a deeper level, collections of model weights---left by fine-tuning tasks---have long served as implicit representations, supporting operations such as averaging for accuracy~\citep{wortsman2022soups}, constructing skill vectors~\citep{ilharco2023task}, and interpolation through training time~\citep{nylund2024time}. Fine-tuned models cluster in connected, low-loss regions~\citep{gueta2023knowledge} so that modular adapters can be merged~\citep{chronopoulou2023adaptersoup,huang2024lorahub,yadav2023ties}. However, these weight-based representations are fundamentally unidentifiable: optimizer randomness erases any canonical correspondence~\citep{garipov2018loss,frankle2020linear,entezari2022permutation,ainsworth2023gitrebasin}, and the coordinates are not stable or comparable across models. Hypernetworks~\citep{ha2017hypernetworks} resolve this by generating adapters in a consistent reference frame. Prior works focused on conditioning these on descriptions or tasks~\citep{ye2021hypter,mahabadi2021hyperformer,phang2023hypertuning,charakorn2025texttolora}. Among them, \texttt{T2L}~\citep{charakorn2025texttolora} is the closest comparison, as it produces a LoRA adapter from text.
Unlike \doctolora, \texttt{T2L} does not reconstruct the original document but instead produces task-specific LoRA adapters. It embeds text via a frozen \texttt{GTE} encoder~\citep{li2023gte}, creating an information bottleneck as the embedding is not optimized for document reconstruction. In contrast, \doctolora conditions directly on the document for reconstruction, making its adapter an identifiable, decodable representation. We include \texttt{T2L} as a baseline (\secref{results}).

\section{Discussion}
\label{sec:discussion}

We showed that the adapter a frozen \doctolora hypernetwork generates from a paper serves as a decodable, semantically meaningful representation of the idea in it.
Using simple arithmetic operations on the embedding, we generated representative labels for clusters and interpolated between ideas.
Our core contribution is conversational embeddings that allow researchers to query underlying ideas directly, moving beyond passive inspection of static vectors.
This capability enables exploring the semantic structure of research areas without established taxonomies.

\doctolora embeddings form a smooth semantic space, where averaging and interpolating produce coherent descriptions (\secref{cluster-labeling}, \secref{fusion}).
We hypothesize that this smoothness arises from mapping documents into a shared coordinate system of generative weights, where each point parameterizes a generative model.
This explanation remains a conjecture based on empirical observations, and formally evaluating how the resulting distributions blend remains open for future work.

There is a tension between search and generation, as illustrated by \texttt{T2L} and \doctolora.
\texttt{T2L} is built on \texttt{GTE} embeddings for similarity search, which retrieves documents well, but decoding from these embeddings fails to reconstruct the source (\secref{cluster-labeling}).
In contrast, \doctolora adapters are trained for generation and capture full document content, but perform poorly on retrieval by cosine similarity (\secref{similarity}).
We bridge this gap with an invertible map $g_\theta$ (\secref{kron-method}), making \doctolora embeddings usable for search without losing decodability.
Trained under the same supervision, this map leaves dedicated text encoders nearly unchanged (Appendix~\ref{app:symmetric}).
The transform therefore surfaces structure already present in the \doctolora embedding rather than introducing new geometry.

The \doctolora embedding is a joint function of the document and the base model.
The hypernetwork generates an adapter on top of a base model with broad prior knowledge.
Content already held by the base model need not be repeated in the adapter.
The embedding therefore represents the document relative to this prior.
This relative encoding emphasizes what a document adds beyond common knowledge.
The prior also fixes the coordinate system.
The same document yields different embeddings under different base models, each prior reshaping the geometry.
Embeddings from different checkpoints do not share a space and cannot be pooled or compared directly.
The choice of base model is part of the representation.

Several limitations remain.
First, we test composition only along edges between two ideas.
Whether the same behavior holds for mixtures of three or more sources is untested.
Second, the embedding depends on its base model as well as the document.
Two base models can therefore order the same documents differently.
Third, the embedding is large and its encoder is larger.
It carries an adapter for every layer, $147{,}456$ coordinates at full rank and $18{,}432$ once pooled, and it is produced by a $4.0$B base model against the $110$M of \texttt{SBERT}.
Whether the tensor can be compressed without losing decodability remains untested.
Current \doctolora architectures generate adapters across all layers, though individual layers may not contribute equally to generation.
Selectively omitting adapters from certain layers might maintain response quality, or hypernetworks could be trained to target only specific layers to reduce parameter counts.
These directions outline architectural refinements for future work.

Conventional embeddings reduce documents to static vectors that cannot render the space between papers as readable text.
\doctolora establishes conversational embeddings where every point expands into an adapter that can be queried in natural language.
We expect the ability to query embeddings to be especially valuable for investigating unexplored regions in the landscape of science, including knowledge holes, interdisciplinary gaps, and frontiers of knowledge.
By enabling direct dialogue with constructed arithmetic points, conversational embeddings offer a path to gain insights into potential new ideas that lie between existing publications.

\section*{AI use statement}
\label{sec:ai-use}

We manually wrote the initial drafts of the manuscript, then refined and reviewed the text using generative AI tools. 
We use AI tools for experiment coding.
For literature search, we use Google Scholar and generative AI tools to help search for relevant literature, with all surfaced references checked by the authors.
All results and outputs were verified against source data, and we take full responsibility for the content of the paper.

\section*{Reproducibility Statement}
\label{sec:repro}

\ificlrpreprint
The code, configurations, and Snakemake workflow are available at \url{https://github.com/skojaku/doc2lora-embedding}.
All experiments can be fully reproduced using this workflow.
\else
All experiments can be fully reproduced using the code, configurations, and Snakemake workflow included with this submission.
\fi
The workflow covers every result and figure in the manuscript. The main results and manuscript figures are generated from the released \texttt{SakanaAI/doc-to-lora} checkpoints. 
Cluster-label comparisons and incoherent-cluster controls are judged by a panel of five language models from different providers, ensuring that no single model or naming convention dominates the scoring. The models are accessed via established APIs such as Google Vertex AI and OpenRouter. Importantly, no model judges its own output, and no label is rewritten by another model, so the scoring process is independent and unbiased.
While the external judge models are versioned proprietary endpoints and may change over time, scoring within a fixed panel is deterministic and reproducible, i.e.,  inputs are presented in a stable order, and responses are cached, ensuring that repeated runs produce identical outcomes. Any label overlap analysis is model-free and fully reproducible from the released outputs.

\ificlrpreprint
\section*{Funding}

Research reported in this publication was supported by SUNY System Administration using the SUNY AI Platform.
This work was also supported by the SUNY-IBM AI Research Alliance (Awards \#96404 and \#96340).

\section*{Author contributions}

S.K. conceived the study, implemented the code, performed the experiments, analyzed the data, and wrote the manuscript.
C.M. implemented the code and performed the experiments.
J.Z. discussed the results and wrote the manuscript.
All authors reviewed and edited the manuscript.
\fi

\bibliography{main}
\bibliographystyle{iclr2027_conference}

\appendix
\section{Datasets}
\label{app:data}

We evaluate on three academic domain corpora and the five S2AND author name disambiguation datasets.
Physics comes from the American Physical Society (APS) corpus.
Economics and Psychology are subsets cut out of OpenAlex (a bulk snapshot from March $2026$) based on journals.
We convert every paper into a single string \texttt{Title: \ldots{} Abstract: \ldots} and embed it once.

\textbf{Physics (APS).}
The APS corpus contains $644{,}022$ papers from $18$ \emph{Physical Review} journals ($1893$--$2020$), $8.3$M citation edges, and author-specified PACS topic codes.
Although the public release omits abstracts, we recover them from OpenAlex via DOI matching.
$99.3\%$ of the APS papers match OpenAlex works, and $97.3\%$ of those contain abstracts.
We embed all $644{,}022$ papers regardless of whether we find an abstract.
Every encoder receives the same string.
Missing abstracts therefore introduce no bias into the comparison.
To label clusters, we use PACS nodes from all three levels of the hierarchy (\secref{cluster-labeling}).
Condensed matter accounts for $42.5\%$ of the corpus.

\textbf{Economics and Psychology (OpenAlex).}
For each discipline, we extract Q1/Q2 journals from the Scimago Journal Rank (SJR) in the target subject areas and match them by ISSN to OpenAlex venues.
Economics uses ``Economics, Econometrics and Finance'' ($1{,}035$ journals), and Psychology uses ``Psychology'' ($1{,}166$ journals).
A paper is included if it appears in a matched venue, has a non-empty scientific-work type and title, and has at least one author with $\ge 2$ papers within the slice.
This final condition excludes one-off author--venue combinations and preserves the connectivity of the authorship network.
For text-based tasks, we also require abstracts of $\ge 300$ characters and primary Scopus/ASJC topic labels.
This filter leaves $565{,}475$ Economics papers and $986{,}800$ Psychology papers.
Economics spans $26$ major fields and $244$ subfields.
Psychology spans $26$ major fields and $245$ subfields.
The Scopus/ASJC labels come from OpenAlex and are independent of the SJR venue rules.

\textbf{Train-evaluation overlap.}
Both the transformation training pool and the evaluation corpus were obtained from OpenAlex. 
We validate the overlap by checking that evaluation papers account for less than $1\%$ of the training set, and training citation edges falling within a single evaluation field constitute no more than $1.24\%$. Since we do not use any author labels for training $g_\theta$, the author-merging supervision is completely separated. To verify the absence of information leakage, we retrained $g_\theta$ using only citations from $2018$ or earlier; the change in next-paper AUC was at most $.005$, and the ranking of our method remained unaffected. Similarly, even when topic classification is split by time (training $\leq$ 2016, testing $>$ 2016), the encoder's rankings are preserved. For Physics, since the labels are journal codes and the benchmark ended in $2015$, we did not perform a temporal split.

\textbf{Author-name disambiguation (S2AND).}
The five subsets of S2AND~\citep{subramanian2021s2and} are zbMATH ($15{,}181$ papers), QIAN ($6{,}542$), ArnetMiner ($7{,}144$), PubMed ($2{,}871$), and KISTI ($40{,}383$).
Each provides ground-truth author clusters for ambiguous name blocks.
We use title and abstract texts where available, leaving name blocks and metadata untouched.

\section{Recipe decoding and fusion example}
\label{app:recipe-fusion}

We illustrate recipe fusion using two short recipe documents—Italian (cacio e pepe) and Japanese (kake udon), each encoded independently by the Mistral-7B \doctolora hypernetwork. 
We decoded the recipes at both endpoints and their interpolation (full-rank mode, \secref{doc2lora-embedding}), by prompting: ``Write out this recipe: give the dish a name, list the ingredients, and describe the preparation steps.'' 
For example, with only the Italian adapter active, the model answers ``The cheese used in this recipe is Pecorino Romano.'' when asked about the cheese, even without seeing the recipe text. 
Mistral-7B was used (rather than Qwen3-4B) for its more natural recipe outputs, but this does not affect the results.
All outputs below are shown verbatim.

\textbf{Source document $A$ (Italian).}

\begin{snugshade}
\begin{Verbatim}[breaklines=true,breakanywhere=true,fontsize=\footnotesize]
Cacio e Pepe (Roman pasta with cheese and black pepper)

A classic recipe of the Roman tradition with only three ingredients. Serves 2.

Ingredients:
- 200 g spaghetti (or tonnarelli)
- 100 g Pecorino Romano cheese, finely grated
- 2 teaspoons whole black peppercorns
- coarse salt for the pasta water

Preparation:
1. Toast the black peppercorns in a dry skillet over medium heat until
   fragrant, then crush them coarsely in a mortar.
2. Bring a pot of lightly salted water to a boil and cook the spaghetti until
   one minute short of al dente. Use less water than usual so the cooking
   water becomes rich in starch.
3. While the pasta cooks, mix the grated Pecorino Romano with a ladle of the
   hot, starchy pasta water, stirring vigorously until it forms a thick,
   smooth cream with no lumps.
4. Transfer the spaghetti to the skillet with the crushed pepper, add a splash
   of pasta water, and finish cooking for one minute, tossing constantly.
5. Off the heat, add the Pecorino cream and toss vigorously, loosening with
   more pasta water as needed, until every strand is coated in a glossy sauce.
6. Serve immediately with extra grated Pecorino and a final grind of pepper.
\end{Verbatim}
\end{snugshade}

\textbf{Source document $B$ (Japanese).}

\begin{snugshade}
\begin{Verbatim}[breaklines=true,breakanywhere=true,fontsize=\footnotesize]
Kake Udon (Japanese udon noodles in hot dashi broth)

A simple and warming Japanese noodle soup built on dashi stock. Serves 2.

Ingredients:
- 2 portions fresh or frozen udon noodles
- 600 ml dashi stock (kombu and katsuobushi)
- 3 tablespoons soy sauce
- 2 tablespoons mirin
- 1 teaspoon sugar
- 2 spring onions, thinly sliced
- shichimi togarashi (Japanese seven-spice), to taste
- optional toppings: kamaboko fish cake, tempura flakes (tenkasu), a soft egg

Preparation:
1. Make the dashi: soak a piece of kombu in cold water for 30 minutes, bring
   it almost to a boil and remove the kombu, then add a handful of
   katsuobushi (dried bonito flakes), turn off the heat, let it steep for
   2 minutes, and strain.
2. Season the dashi with the soy sauce, mirin, and sugar, and keep the broth
   hot over low heat without boiling.
3. Cook the udon noodles in a separate pot of unsalted boiling water according
   to the package directions, then drain and rinse briefly to remove surface
   starch.
4. Divide the noodles between two deep bowls and ladle the hot seasoned broth
   over them.
5. Top with sliced spring onions and any optional toppings, and finish with a
   pinch of shichimi togarashi. Serve immediately, very hot.
\end{Verbatim}
\end{snugshade}

\textbf{Decoded midpoint, $\alpha=0.5$.}

\begin{snugshade}
\begin{Verbatim}[breaklines=true,breakanywhere=true,fontsize=\footnotesize]
Recipe Name: Kake Udon with Creamy Parmesan Sauce

Ingredients:
- 200 g fresh udon noodles
- 2 tablespoons unsalted butter
- 2 tablespoons mirin
- 2 tablespoons soy sauce
- 1 teaspoon wasabi paste
- 1/2 cup grated Parmesan cheese
- 1/2 cup heavy cream
- 1 spring onion, thinly sliced
- Optional: a pinch of freshly ground black pepper

Preparation Steps:
1. Cook the fresh udon noodles according to the package instructions. Drain
   and rinse with cold water.
2. In a deep skillet or pan, melt the unsalted butter over medium heat.
3. Add the mirin and cook fsr 1 minute.
4. Stir in the soy sauce and wasabi paste.
5. Gradually add the grated Parmesan cheese, stirring constantly to prevent
   lumps from forming.
6. Pour in the heavy cream and bring to a simmer.
7. Cook for 2 minutes, stirring occasionally, until the sauce has thickened.
8. Add the cooked udon noodles to the skillet and toss to coat them evenly
   with the sauce.
9. Remove from heat and top with the sliced spring onion.
10. Optional: finish with a pinch of freshly ground black pepper.
11. Serve immediately.
\end{Verbatim}
\end{snugshade}

The midpoint describes a dish that appears in neither source. Udon noodles coated in a melted-cheese cream sauce, the cacio e pepe technique, seasoned with the mirin and soy of the Japanese broth and finished with the spring onion of one source and the black pepper of the other. The endpoints are faithful. $\alpha=0$ decodes to cacio e pepe and $\alpha=1$ to kake udon, each with minor paraphrases. Off-midpoint interpolants snap to the nearer source ($\alpha=0.25$ decodes to cacio e pepe; $\alpha=0.75$ to kake udon with butter entering the broth).
We note that the blend is imperfect in an instructive way. Namely, the midpoint substitutes Parmesan for Pecorino Romano and introduces heavy cream found in neither source, so the fusion is a plausible blend of the two cuisines rather than a constraint-satisfying merge of the two ingredient lists.

\section{Bijective transforms}
\label{app:bijective}

The transform $g_\theta$ must be invertible so that we can always recover the adapted representation needed for decoding (\secref{kron-method}). We guarantee this invertibility by structuring the channel transformation matrix $\matrix{C}$. Specifically, we write $\matrix{C}$ as
\begin{align}
   \label{eq:c-factor}
   \matrix{C} = \matrix{R} \, \mathrm{diag}(\sigma_1, \ldots, \sigma_{n_d}),
   \qquad \sigma_1, \ldots, \sigma_{n_d} > 0,
\end{align}
where $\matrix{R}$ is a rotation matrix (orthogonal) and $\mathrm{diag}(\sigma_1,\ldots,\sigma_{n_d})$ is a diagonal matrix with all positive entries. Both of these components are invertible, i.e., a diagonal matrix with positive values is inverted simply by taking the reciprocal of each entry, and a rotation is inverted by transposing it. Therefore, their product $\matrix{C}$ is also invertible.

To make sure that $\matrix{R}$ stays a true rotation during training, we construct it by exponentiating a skew-symmetric matrix:
\begin{align}
   \label{eq:q-expm}
   \matrix{R} = \exp(\matrix{P} - \matrix{P}^{\top})
\end{align}
where $\matrix{P}$ is any square matrix of size $n_d \times n_d$ (with no constraints). The difference $\matrix{P} - \matrix{P}^{\top}$ is always skew-symmetric by construction, and exponentiating any skew-symmetric matrix produces an orthogonal (rotation) matrix~\citep{lezcano2019cheap}, which lets us freely optimize $\matrix{P}$ as unconstrained parameters.

The full inverse of $g_\theta$ is given by
\begin{align}
   \label{eq:transform-inverse}
   g_\theta^{-1}(\matrix{V}') = a_\ell^{-1} \matrix{V}' \matrix{C}^{-1},
\end{align}
where $a_\ell > 0$ is a scalar learned separately for each layer (\secref{kron-method}). Since $g_\theta$ is linear, operates along the channel axis, and uses a simple per-layer scaling, it commutes with mean-pooling over the rank axis.
In other words, it produces the same result if we pool the rank dimension before or after applying $g_\theta$. Therefore, we can use $g_\theta$, trained on full-rank tensors, on mean-pooled vectors as well.

We initialize $g_\theta$ to the identity (i.e., no transformation initially) and keep all parameters of the base language model frozen. Training for $g_\theta$ uses $42{,}332$ examples from OpenAlex (see Appendix~\ref{app:data}) paired up as anchor, positive, easy negative, and hard negative.
The easy and hard negatives follow the sampling protocol of \texttt{SPECTER}. An easy negative is a paper drawn uniformly at random from the corpus, excluding the anchor and the papers it cites. A hard negative is a paper cited by the positive but not by the anchor, i.e., at distance two from the anchor. The optimization uses an InfoNCE contrastive loss with temperature $0.05$, in which each anchor is scored against its own hard and easy negative together with the other $255$ positives in the batch. We train for $8{,}000$ Adam steps with batch size $256$ at a learning rate of $10^{-3}$ and no added regularization.

\section{Cluster-label evaluation protocol}
\label{app:baseline-labels}

We evaluate labeling over a 28-node PACS hierarchy slice (four fields, each with two categories and two subcategories). 
All methods are given the same cluster members and asked to assign scientific field labels. Compared methods include \doctolora, \texttt{ICAE}, \texttt{KeyLLM}, \texttt{BERTopic}, \texttt{vec2text}, \texttt{T2L}, and a baseline (\texttt{in-context}) using only the base Qwen3-4B LLM prompted with abstracts from the 20 members nearest each centroid. \texttt{ActPatch} (hidden-state baseline) is evaluated separately in Appendix~\ref{app:actpatch}.

\doctolora receives its instruction with the cluster documents internalized in the adapter and no document text in the prompt:
\begin{quote}\itshape
In 2 to 3 words, name the scientific field that all of these documents belong to.
Reply with only the field name.
\end{quote}
\texttt{ICAE} receives the same prompt verbatim and decodes against its averaged memory slots.
\texttt{T2L} receives the same prompt and decodes using its training prompt format.
\texttt{T2L} turns a document into an adapter in three stages (a \texttt{GTE} vector, a $64$-dimensional TaskEncoder output, and the LoRA weights), so a cluster could be averaged at any of them.
We average the TaskEncoder outputs, as in the mixing comparison (Appendix~\ref{app:edge-protocol}); averaging at the other two stages yields similar unrelated field names (``Social Sciences'', ``Information science'').
\texttt{KeyLLM}, \texttt{BERTopic}, and \texttt{vec2text} cannot take this instruction.
\texttt{KeyLLM} is a keyword extractor returning a ranked list of keywords.
\texttt{BERTopic} represents a cluster by its ten most distinctive words (Appendix~\ref{app:bertopic}).
\texttt{vec2text} is an iterative inverter trained to reconstruct text from an embedding and has no prompt interface.
We therefore score every method on its native output without rewriting any label through a second language model.
Passing outputs through an auxiliary namer would test the namer.
The comparison asks whether the representation itself supports the abstraction.

\begin{table}[h]
    \centering
    \scriptsize
    \setlength{\tabcolsep}{4pt}
    \caption{Cluster-labeling protocols comparing embedding-arithmetic (top) and text-based (bottom) methods without downstream language model post-processing: ``Operates on'' denotes the representation averaged and labeled, ``\#docs'' indicates consumed member documents per node, and ``Native output'' defines directly evaluated raw outputs.}
    \label{tab:label-protocols}
    \begin{tabular}{@{}l l l l@{}}
    \toprule
    Method & Operates on & \#docs & Native output (scored as-is) \\
    \midrule
    \doctolora & embedding (mean adapter) & up to $2{,}000$ & field name (2--3 words) \\
    \texttt{ICAE} & embedding (mean memory slots) & $8$ & field name (2--3 words) \\
    \texttt{T2L} & embedding (mean TaskEncoder output) & up to $2{,}000$ & field name (2--3 words) \\
    \texttt{vec2text} & embedding (mean GTR vector, inverted) & $20$ & reconstruction (no prompt) \\
    \midrule
    \texttt{KeyLLM} & text (medoid document) & $1$ & keyword list \\
    \texttt{BERTopic} & text (members, c-TF-IDF) & up to $2{,}000$ & keyword list \\
    \texttt{in-context} & text (centroid-nearest abstracts) & $20$ & field name (2--3 words) \\
    \bottomrule
    \end{tabular}
\end{table}

This design categorizes the methods based on output length. Compared to the official PACS labels of $4.0$ words, both \doctolora and \texttt{ICAE} respond to prompts with $2.4$ words each. Therefore, their scores are directly comparable. \texttt{KeyLLM} averages $8.8$ words, \texttt{BERTopic} $10$ words, and \texttt{vec2text} $15.5$ words. The inability to adhere to a concise naming format is an inherent characteristic of the methods, and we do not apply length corrections. \texttt{T2L} outputs average $2.0$ words. Across all methods, \doctolora, \texttt{ICAE}, \texttt{T2L}, and \texttt{vec2text} operate exclusively on embeddings, whereas \texttt{KeyLLM}, \texttt{BERTopic}, and the \texttt{in-context} read the member text. Among the embedding methods, \doctolora differs in that it reads the abstraction level from the length of the mean adapter vector (Appendix~\ref{app:length-dial}).

\textbf{Fuzzy overlap} compares native outputs directly against official PACS labels using a deterministic token-set ratio in \texttt{rapidfuzz}.
The score ranges from $0$ to $1$, involves no language model, and reproduces deterministically from the released strings.
Because fuzzy overlap measures surface tokens, verbose outputs receive lower scores against four-word reference labels, and the metric cannot credit valid paraphrases.

The \textbf{pairwise win rate} evalutes each method using five language-model judges (Reproducibility Statement, p.~\pageref{sec:repro}). For each cluster, judges compare the official PACS label and two candidate method labels, picking the closer one in meaning or declaring a tie; all pairs are shown in both label orders to remove position bias, i.e., a win counts only if both orders agree, otherwise scored as a tie ($0.5$). 
The seven methods of \tabref{label-eval} enter the round robin, each compared with every other on every node. A method's win rate is the fraction of its comparisons that it wins, averaged over nodes. \figref{cluster-labels}b shows the rate of each row method against each column method, and its row mean is the win rate. 
Judges rate \doctolora labels as tied with official PACS names on $32\%$ of judge-node decisions.
By comparison, this rate is $25\%$ for \texttt{ICAE}, $22\%$ for the \texttt{in-context}, and $0\%$ for \texttt{KeyLLM}, \texttt{BERTopic}, and \texttt{vec2text}.

\begin{table}[h]
\centering
\small
\caption{
    Cluster-label quality over $28$ PACS nodes. Each method's native output is scored against the true node label. \textbf{Fuzzy overlap:} deterministic token-set ratio with the reference. \textbf{Pairwise:} win rate from a five-judge round robin over the seven methods below; ties score $0.5$ and require both label orders to agree. $\pm$ is the bootstrap s.d.\ over $1{,}000$ node resamples. \textbf{Words:} mean label length (\textit{PACS:} $4.0$). The ground-truth control has no win rate. Best (excluding control) in bold. Full pairwise matrix: \figref{cluster-labels}b.
}
\label{tab:label-eval}
\setlength{\tabcolsep}{5pt}
\begin{tabular}{@{}l c c c@{}}
\toprule
Method & Words & Fuzzy overlap & Pairwise \\
\midrule
\textit{Ground truth (control)} & $4.0$ & $1.000$ & --- \\
\midrule
\doctolora & $2.4$ & $\mathbf{.642 \pm .055}$ & $\mathbf{.755 \pm .032}$ \\
\texttt{ICAE} & $2.4$ & $.573 \pm .058$ & $.678 \pm .039$ \\
\texttt{KeyLLM} & $8.8$ & $.311 \pm .019$ & $.474 \pm .034$ \\
\texttt{vec2text} & $15.5$ & $.302 \pm .026$ & $.215 \pm .021$ \\
\texttt{BERTopic} (top $10$) & $10.0$ & $.296 \pm .020$ & $.696 \pm .033$ \\
\texttt{T2L} & $2.0$ & $.277 \pm .010$ & $.002 \pm .001$ \\
\midrule
\texttt{in-context} & $2.4$ & $.499 \pm .051$ & $.678 \pm .031$ \\
\bottomrule
\end{tabular}
\end{table}

\doctolora achieved the highest scores on both metrics, with a fuzzy overlap of $.642 \pm .055$ and a pairwise win rate of $.755 \pm .032$ (\tabref{label-eval}).
Two of the five judges rank \doctolora first by win rate and three rank \texttt{BERTopic} first, because \texttt{BERTopic} beats the weakest methods by wider margins (e.g., $.850$ against \texttt{KeyLLM}, where \doctolora has $.750$).
Head to head, the judges still prefer \doctolora over \texttt{BERTopic} ($.604$; Appendix~\ref{app:bertopic}).
\doctolora’s direct baseline is \texttt{ICAE}, which processes identical instructions and generates outputs of the same average length.
\doctolora outperforms \texttt{ICAE} in pairwise win rates ($.755$ vs. $.678$).
Since their marginal intervals overlap in terms of fuzzy overlap ($.642 \pm .055$ vs. $.573 \pm .058$), we examine the paired node-level differences.
The mean paired difference is $+.069$, the bootstrap standard deviation is $.034$, and the $95\%$ confidence interval is $[.016, .145]$.
Out of a total of $28$ nodes, \doctolora achieved a higher overlap at $12$ nodes, \texttt{ICAE} achieved it at $1$ node, and $15$ nodes were tied.
In direct head-to-head comparisons, \doctolora and \texttt{ICAE} tied in $103$ out of $140$ decisions; \doctolora won approximately two-thirds of the remaining decisions, resulting in a head-to-head win rate of $.596$.
Compared to \texttt{in-context} (pairwise $.678$, fuzzy $.499$), \doctolora recorded a head-to-head win rate of $.629$, with $78$ ties.
Therefore, decoding an averaged adapter recovers the field at least as well as reading the member abstracts.

\doctolora is the top performer on both metrics and wins every head-to-head comparison. 
Compared to its closest baseline, \texttt{ICAE}, which operates under the same conditions, \doctolora leads in win rate, and their other scores are similar enough to warrant direct comparison. Analysis at the node level shows \doctolora typically outperforms or ties with \texttt{ICAE}. 
In head-to-head matchups, \doctolora has a clear edge, and it also surpasses the \texttt{in-context} approach. 
Overall, decoding an averaged adapter is at least as effective as reading member abstracts.

\texttt{T2L} performs worst on both metrics, outputting only two field names (``Social Sciences'' and ``Social psychology'') across all $28$ nodes, with a fuzzy overlap floor of $.277$. It loses almost every pairwise comparison, including against \texttt{vec2text}. The method's mapping does not yield decodable representations. The synthesized adapters expanded from the frozen \texttt{GTE} encoder cannot generate meaningful labels. We note that cluster-labeling is not the intended use case and goal for \texttt{T2L}.

Native outputs for every PACS node appear in \tabref{hierarchy-labels} in Appendix~\ref{app:tables}.
\doctolora outputs match official designations, producing ``Quantum chromodynamics'' for code 12.38 and ``Fluid dynamics'' for code 47.
\texttt{KeyLLM} reflects single-document specificity, returning ``square-lattice Heisenberg antiferromagnet, S=1/2, nearest-neighbor exchange'' for magnetic ordering (75.10).
\texttt{ICAE} identifies ``Condensed matter physics'' for code 71 and ``Nonlinear optics'' for code 42.65, but occasionally shifts hierarchy levels, returning ``Materials science'' for bulk band structure (71.20) and ``Quantum many-body systems'' for general physics (0).
\texttt{vec2text} reconstructions are garbled, yielding ``in-distance interactions between electron-bonding spectral properties'' for code 3 and ``inferromagnetic spin-diabetic scattering interactions'' for code 75.

\subsection{\texttt{BERTopic} baseline}
\label{app:bertopic}

The standard \texttt{BERTopic} pipeline finds clusters (UMAP + HDBSCAN) and then uses c-TF-IDF to extract words that are frequent within a cluster and rare elsewhere; as our evaluation assumes clusters are fixed, we run BERTopic in its manual topic modeling mode (bypassing clustering via empty \texttt{BaseDimensionalityReduction} and \texttt{BaseCluster}, with PACS nodes as clusters) to test only the labeling step. We match \doctolora's setup by providing identical titles, abstracts, a maximum of $2{,}000$ papers per node (seed $0$), and precomputed SBERT embeddings. Since c-TF-IDF compares each cluster to all others, output depends on the full set of clusters. Therefore, we fit a model for each PACS hierarchy level, including all groups at that level with at least $100$ papers: $9$ fields ($18\text{k}$ papers), $65$ divisions ($99\text{k}$), and $393$ subdivisions ($305\text{k}$), not just the nodes we score.

We evaluate \texttt{BERTopic} using its default top-10 keyword list as the label, and also try a shorter top-3 list to control for label length, since longer labels are penalized by token-set metrics. While \texttt{BERTopic}'s fuzzy overlap score is $.296$ for the top-10 (or $.346$ for the top-3), similar to \texttt{KeyLLM} ($.311$) and \texttt{vec2text} ($.302$), the judges tend to group \texttt{BERTopic} with methods that successfully name the field. In direct comparisons, \texttt{BERTopic} performs worse than \doctolora (winning only $.396$ of matchups), about the same as \texttt{in-context} ($.493$), close to \texttt{ICAE} ($.471$), and better than \texttt{KeyLLM} ($.850$) and \texttt{vec2text} ($.964$). This suggests that fuzzy overlap in keyword-based labels is influenced more by length and wording than by content. Although using the top-3 list improves fuzzy overlap by $.05$, judges only prefer it over the top-10 list in $17.5\%$ of cases.

\texttt{BERTopic} never outputs the field name itself; instead, it produces lists of common subtopics found within each field, like ``quark, mass, model, gauge, higgs, theory, qcd, decays, decay, production'' for Elementary particles (node 6), or ``quantum, time, state, systems, states, model, entanglement, black, field, phase'' for General (node 0). Because few words are unique to a field, these lists cannot represent the field name clearly. Judges never considered BERTopic's label equivalent to the correct field name ($0\%$ of cases, both for the regular and top-3 lists), whereas methods like \doctolora, \texttt{ICAE}, and \texttt{in-context} tied with the correct name $32\%$, $25\%$, and $22\%$ of the time, respectively. The top-3 word list was evaluated separately from the main results, always using the same judges, protocol, and random seed. Finally, we did not include results from the standard BERTopic pipeline, as it produces its own clusters instead of labeling the original PACS nodes.

\subsection{Validity of cluster labels}
\label{app:incoherent}

We test whether decoded cluster labels require a coherent member set (\tabref{incoherent}).
For each of $30$ real PACS nodes, we construct two control clusters of identical size by sampling papers uniformly across PACS chapters outside the target node.
We decode one control at its native centroid length of $11.15$ and a second control scaled to the real-node mean of $11.49$.

A five-judge panel sees each label alongside twelve sampled member titles and is not told which method or cluster type a row comes from.
The panel accepts the real-node label on $73.3\%$ of clusters and rejects $3.3\%$ (one of $30$).
The remainder are judged partial.
Judges accept zero control labels, rejecting $100\%$ of the native control and $96.7\%$ of the norm-matched control.
Control decodings collapse almost uniformly to ``Quantum field theory''.
The norm-matched control earns the same verdict as the native control.

\begin{table}[h]
\centering
\footnotesize
\caption{Scores from the blinded five-judge panel on the incoherent-cluster control experiment. \emph{PACS node (real)} denotes actual clusters, while the control rows consist of clusters of the same size sampled across different PACS chapters and decoded either at their native centroid norm or rescaled to the real nodes' norm. \emph{belongs=yes} and \emph{rejected} indicate the proportion of clusters whose label the panel accepted as valid for the members or rejected entirely. The rest were judged partial.}
\label{tab:incoherent}
\begin{tabular}{lrrr}
\toprule
cluster set & $n$ & belongs=yes & rejected\\
\midrule
PACS node (real) & 30 & 73.3\% & 3.3\% \\
cross-chapter control & 60 & 0\% & 100\% \\
cross-chapter, norm-matched & 60 & 0\% & 96.7\% \\
\bottomrule
\end{tabular}

\end{table}

\subsection{Adapter length sets label granularity}
\label{app:length-dial}

Vector length governs the level of abstraction in decoded \doctolora embeddings.
\secref{cluster-labeling} shows that scaling an embedding by $\alpha$ shifts the decoded label from specific topics to general fields (for instance, moving from citric acid cycle to cellular respiration) until near the origin the label reverts to the prior distribution of the prompt.
We evaluate this magnitude sweep on \texttt{ICAE} and \texttt{vec2text} across five seed documents covering a physics concept, a biochemistry process, a fruit, a historical person, and a patent.
\figref{cluster-labels}d shows the \doctolora sweep on four of these seeds.
The person seed appears only in the tables of Appendix~\ref{app:tables}.

Neither baseline exhibits upward semantic abstraction under vector scaling.
\texttt{ICAE} remains semantically unchanged across lengths, returning the specific concept label at every value of $\alpha$ with variations limited to capitalization and surface wording.
\texttt{vec2text} maintains surface-level specificity while the reconstruction remains coherent, then decomposes into unrelated text fragments at low $\alpha$ without naming broader fields (\tabref{radial-vec2text}).
Among the three embedding methods on this sweep, the specific-to-general ladder is specific to \doctolora.

\section{Activation patching does not recover the source document}
\label{app:actpatch}

LatentQA~\citep{pan2024latentqa} could not be reproduced from the released code and checkpoint.
Intended as a trained activation decoder baseline, it yields no reported metrics.
The claim we could have drawn is that a trained activation decoder also fails, and a broken baseline produces exactly the appearance our result wants.
We instead report a training-free readout of Qwen3-4B-Instruct hidden states via activation patching~\citep{ghandeharioun2024patchscopes,chen2024selfie}.

For each document, we mean-pool a hidden state $h$ from a chosen layer, insert $h$ into a fixed slot of the completion prompt, and generate tokens.
We sweep layers, slot counts, and decoding style.
The question prompt
\begin{quote}\itshape
Document: ?\\[0.5em]
In one sentence, what is the specific topic of this document?
\end{quote}
reports that the document is empty.
The continuation prompt
\begin{quote}\itshape
The following is a scientific abstract.
\end{quote}
carries topic information.
The sweep selects layer $15$ with eight slots.
This baseline is \texttt{ActPatch}.
Both \doctolora and \texttt{ActPatch} inject a vector into the same model and prompt, differing only in vector origin.
Generated outputs are compared to those from the unpatched prompt to control for prompt prior.

We first test whether the resulting embeddings can identify source documents.
Tested on $500$ abstracts per method retrieving each among $99$ same-subfield distractors, \doctolora retrieved its source $91.8\% \pm 1.2$ (MRR $0.946 \pm 0.008$), while \texttt{ActPatch} managed just $6.0\% \pm 1.1$ (MRR $0.132 \pm 0.011$), which is near chance ($1.0\%$). 
Decodes from activations retain weak document signal. 
Their SBERT similarity to the source is only $0.211$ above baseline ($0.794$ for \doctolora). 
\texttt{ActPatch} decodes also often collapse into repetition ($100$ of $500$, vs $1$ for \doctolora).

We interpolate the same $100$ document pairs as \secref{fusion} by decoding three points along each interpolant ($\alpha=0$, $0.5$, $1$), using a linear mix of full \doctolora embeddings or mean-pooled hidden states for \texttt{ActPatch}.
Each decode $D$ of papers $A$ and $B$ is scored by how much its SBERT similarity (all-mpnet-base-v2, see Appendix~\ref{app:edge-protocol}) to $A$ and $B$ exceeds the baseline similarity between $A$ and $B$.
\begin{equation}
\gamma(D;A,B)=\min\bigl(\mathrm{sim}(D,A),\,\mathrm{sim}(D,B)\bigr)-\mathrm{sim}(A,B).
\label{eq:midpoint-gain}
\end{equation}
A verbatim copy of $A$ or $B$ scores $\gamma=0$.
$\gamma>0$ indicates $D$ is closer to both sources than they are to each other.
$\gamma<0$ means at least one similarity falls below the baseline.

The \doctolora midpoint achieves $\gamma = .171 \pm .011$, positive for 99 of 100 pairs, with endpoints at $-0.020$ and $-0.008$.
The \texttt{ActPatch} midpoint yields $\gamma = -0.039 \pm .015$.
Out of the 50 closest pairs, where the sources are most similar and the required threshold for positivity is highest, \doctolora remains positive for all, while \texttt{ActPatch} does so for 8.

Across 28 PACS node means, \doctolora reaches fuzzy overlap $0.575 \pm 0.043$ while \texttt{ActPatch} gets $0.476 \pm 0.038$. Judge agreement is $0.650 \pm 0.068$ for \doctolora and $0.350 \pm 0.065$ for \texttt{ActPatch}. The \doctolora label is closer on 15 nodes, the activation label wins on 5, and there are 8 ties. 
We note that our results do not eliminate the possibility of hidden state-based decoding strategies.
In fact, we note that the \texttt{ActPatch} is train-free and also small in size compared to \doctolora. 
We instead clarify that \doctolora provides values beyond bare hidden-state-based decoding. 

\section{Fusion protocol and metrics}
\label{app:edge-protocol}

We sample paper pairs from the APS corpus, grouped by PACS taxonomy distance. The near group (L1) has pairs within the same subdivision, the far group (L5) has pairs from different subtopics. For each group, we draw 50 pairs and label each paper A or B. \figref{cluster-labels}e and \figref{cluster-labels}f show the near group.
For each pair, we evaluate all methods on 13 interpolation weights from 0 to 1 in steps of 1/12, representing how much paper B contributes. Each method generates outputs at these 13 points using its representation space and the prompt in \secref{fusion}.
\doctolora interpolates the full-rank embeddings of A and B, decoding with Qwen3-4B. \texttt{ICAE} combines memory slots with matching weights then decodes with its native decoder. \texttt{T2L} interpolates TaskEncoder outputs (64 dimensions per document) then runs the hypernetwork to make a LoRA adapter. If we interpolate instead at the frozen \texttt{GTE} input or at the final LoRA factors, the curve stays flat. For the in-context baseline, Qwen3-4B receives both document texts and the mixing percentages directly.
These explicit percentages give the in-context baseline mixing weights that are not available to the other methods.

\begin{quote}
\begin{Verbatim}[breaklines=true,breakanywhere=true,fontsize=\footnotesize]
You are given two short documents:

[A] <corner A>

[B] <corner B>

Imagine ONE research topic that blends these two in the proportions
A:<pa>%
describing this research topic: its problem, methods, and findings.
\end{Verbatim}
\end{quote}

We measure where decoded abstracts land on the axis between the two source papers in SBERT space using all-mpnet-base-v2.
Let $\mathbf{a}$, $\mathbf{b}$, and $\mathbf{d}$ denote the embeddings of paper A, paper B, and the decoded abstract.
The position $t$ is the least-squares projection $t = (\mathbf{d}-\mathbf{a})\cdot(\mathbf{b}-\mathbf{a}) / \lVert \mathbf{b}-\mathbf{a}\rVert^2$.
Under this metric, $t=0$ corresponds to paper A, and $t=1$ corresponds to paper B.
Values of $t$ can fall outside the unit interval $[0,1]$ and are not clipped.
\figref{cluster-labels}e plots the mean position over pairs at each grid point with its $95\%$ bootstrap interval.

We measure verbatim copying as the fraction of word tokens in the decoded abstract that appear identically in either source abstract.
Word tokens are lowercase alphabetic character sequences of length three or greater.
Abstracts randomly sampled from the corpus share roughly $30\%$ of their tokens by chance.
We display this floor as a dashed line in \figref{cluster-labels}f.

\section{Similarity-benchmark protocol}
\label{app:benchmarks}

The field-level benchmarks cover Economics, Psychology, and Physics, with data described in Appendix~\ref{app:data}.
For \doctolora, we evaluate embeddings produced by the Qwen3-4B checkpoint across the primary benchmark, including all three field tasks and the five S2AND datasets.
For retrieval, we mean-pool the adapter tensors across the rank dimension as defined in \secref{doc2lora-embedding}.
Because the transformation $g_\theta$ is linear, pooling before or after applying $g_\theta$ yields identical vectors.
Text baselines comprise \texttt{SPECTER2}~\citep{singh2023scirepeval}, \texttt{SBERT} (all-mpnet-base-v2)~\citep{reimers2019sbert}, \texttt{Instructor} (instructor-large)~\citep{su2023instructor}, \texttt{EmbeddingGemma} (\texttt{embeddinggemma-300m})~\citep{embeddinggemma2025}, and \texttt{GTE} (\texttt{gte-large-en-v1.5})~\citep{li2023gte} used by \texttt{T2L}.
All text encoders run with default settings without task-specific tuning.
\tabref{similarity} lists the absolute scores, and the mean ranks cited in \secref{similarity} are computed from that table.

For the S2AND datasets, we evaluate against the earlier \texttt{SPECTER} model because the S2AND system is designed around those representations.
The benchmark does not provide \texttt{SPECTER2} vectors.
Generating vectors locally would require changing both the text preprocessing and the encoder, which causes the evaluation to diverge from the published benchmark.
Because official vectors do not exist for the other text baselines, we compute their embeddings ourselves from the published titles and abstracts.

For next-paper prediction, papers are split per author over time.
Each query paper is paired with the next paper written by the same author, which forms a positive pair. 
We also pair the query paper with a randomly sampled paper that the author did not write, which forms a negative pair.
For topic classification, we assign OpenAlex papers to one of 26 Scopus or ASJC classes, APS papers to one of 18 Physical Review codes, and sample 60000 papers seen by all encoders, splitting into 80 percent training and 20 percent test, not based on year.
Collaboration prediction uses anchor years 2000 and 2008 for APS and 2008, 2012, and 2016 for Economics and Psychology.
At each anchor, an author is represented by averaging their papers from the previous four years.
Candidate pairs are author pairs at network distance two who share at least one coauthor and have no joint publication.
A positive example is a candidate pair who coauthor a paper together in the next three years after the anchor, while a negative is a pair who do not coauthor in that period.
Because positive and negative pairs have identical local coauthorship structure, only content similarity can distinguish them.
We pool the candidate pairs of all anchor windows and compute a single AUC-ROC over the pooled pairs.
The AUC-ROC of each window is reported in \tabref{collab-per-window}.
For author-name disambiguation, we use the five S2AND subsets from Appendix~\ref{app:data}; each signature is represented by the embedding of its paper.
Signatures are clustered by agglomerative average linkage with a cosine-distance threshold tuned on training data and performance is reported using B3 F1.

Uncertainties are computed via 1{,}000 bootstrap resamples of per-unit evaluation scores; the $\pm$ in \tabref{similarity} denotes the standard deviation across a method's own resamples, with each method resampled independently (so intervals are conservative for comparisons). Mean ranks in \secref{similarity} average over all fourteen benchmarks (not raw scores) to account for differing metrics and scales. Resampling units: next-paper prediction resamples clustered queries; collaboration prediction, candidate author pairs; topic classification, test papers; disambiguation, signatures within name blocks, with B\textsuperscript{3} precision and recall drawn together to compute F1.

Relative model rankings in \tabref{similarity} remain stable across all three base models (\tabref{encoder-matrix} in Appendix~\ref{app:tables}).
The smallest model, Gemma-2-2b, trails Qwen3-4B by up to ${\sim}.05$ on the field tasks but exceeds it on Physics collaboration ($.823$ against $.795$).
Mistral-7B performs comparably to Qwen3-4B and outperforms it on Physics next-paper prediction ($.965$).
Each cell in \tabref{encoder-matrix} reports raw and +$g_\theta$ scores using the evaluation pools, metrics, and bootstrap procedures from \tabref{similarity}, omitting standard deviations for readability.
Because text baseline scores are independent of the base model, we report them only in \tabref{similarity}.

\section{Prompt sensitivity}
\label{app:prompt-sensitivity}

Paraphrasing the decode prompt changes the wording but rarely the content (\tabref{psens}).
For PACS cluster labels and two-paper midpoint descriptions, we decoded each point under eight semantically equivalent prompts, with generation settings held fixed.
We score the eight outputs of a point by SBERT cosine.
Two different PACS nodes decoded under one prompt have mean cosine $.463$, and two different midpoints have mean cosine $.244$.

The eight PACS-label prompts follow (1 is the wording of \secref{cluster-labeling}).
\begin{enumerate}[leftmargin=1.8em,itemsep=2pt,parsep=0pt,topsep=4pt]
\footnotesize
\item In 2 to 3 words, name the scientific field that all of these documents belong to. Reply with only the field name.
\item Using 2 or 3 words, state the scientific field shared by all of these documents. Answer with the field name only.
\item What scientific field do all of these documents belong to? Answer in 2-3 words, nothing else.
\item Name, in at most three words, the research field common to these documents. Output only the name.
\item Give the shared scientific field of these documents as a 2-3 word phrase and nothing more.
\item Identify the field of science that covers all of these documents. Reply with a two or three word field name only.
\item In a phrase of two to three words, what is the common scientific field of these documents? Reply with the phrase alone.
\item Summarise the scientific field of these documents in 2-3 words. Do not add any explanation.
\end{enumerate}

The eight midpoint-description prompts follow (1 is the $2$--$3$ sentence combined-idea prompt).
\begin{enumerate}[leftmargin=1.8em,itemsep=2pt,parsep=0pt,topsep=4pt]
\footnotesize
\item You have read two research papers and hold a single combined idea. In 2 to 3 sentences, describe the combined research idea. Reply with the description only.
\item Describe, in two or three sentences, the single research idea that combines the two papers you have read. Output only the description.
\item What research idea sits between the two papers you have read? Answer in 2-3 sentences.
\item In at most three sentences, state the merged research idea formed from the two papers.
\item Summarise the blended research idea of the two internalised papers in two to three sentences.
\item Give a 2-3 sentence description of the research idea that fuses the two papers you hold.
\item Express, in two or three sentences, the combination of the two papers as one research idea.
\item Write two or three sentences describing the single idea that unites the two papers you read.
\end{enumerate}

PACS labels mostly stay the same string.
Eight prompts produce $3.2$ distinct labels on average.
The most frequent label appears in $72\%$ of the outputs, and $51\%$ of prompt pairs match exactly.

Midpoint descriptions produce a new string under every prompt.
Mean pairwise cosine is $.705$.
The lowest pair is $.507$, still well above the $.244$ baseline between different points.
Two descriptions of the same midpoint therefore remain closer than descriptions of two different midpoints, even when the prompt changes.

The wording in the paper is typical of this paraphrase set.
Similarity to that wording equals or exceeds the mean pairwise similarity ($.841$ for labels, $.705$ for midpoints).

\begin{table}[h]
\centering
\scriptsize
\setlength{\tabcolsep}{4pt}
\caption{Prompt sensitivity. $K$ is the number of semantically equivalent prompts per point, the first being the one the paper reports. \emph{mean pairwise sim.}\ averages SBERT cosine over prompt pairs for the same point; \emph{sim.\ to the reported prompt} compares the other seven against it. \emph{identical strings} is the fraction of those prompt pairs that emit the same text after lowercasing. Two different PACS nodes decoded under one prompt have mean cosine $.463$. Two different midpoints decoded under one prompt have mean cosine $.244$.}
\label{tab:psens}
\begin{tabular}{lrrrrr}
\toprule
decode family & units & $K$ & mean pairwise sim. & sim.\ to the reported prompt & identical strings \\
\midrule
PACS node labels (2--3 words) & 30 & 8 & 0.787 & 0.841 & 51\% \\
two-paper midpoint descriptions & 40 & 8 & 0.705 & 0.705 & 0\% \\
\bottomrule
\end{tabular}

\end{table}

\begin{figure}[t]
\centering
\includegraphics[width=\linewidth]{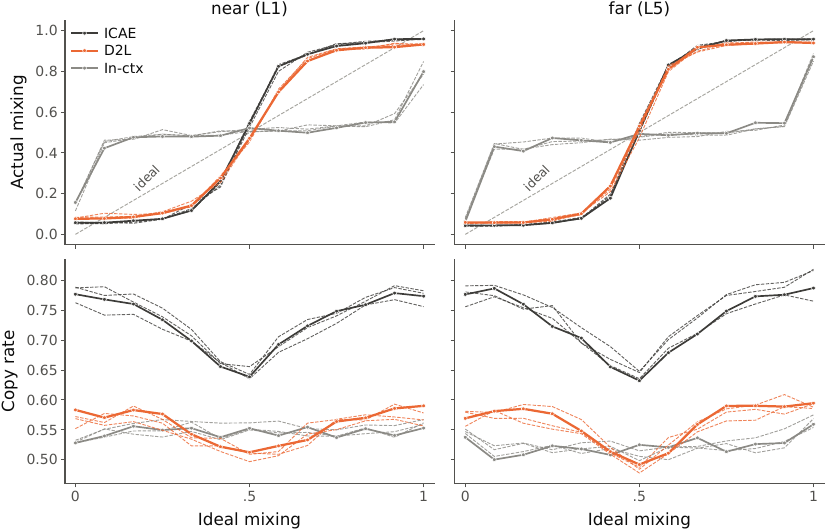}
\caption{Decoded abstracts along the A--B interpolation under paraphrase, near pairs (left) and far pairs (right). Solid lines are the wording reported in \secref{fusion} and dashed lines its three paraphrases. \emph{Top}, where the decoded abstract lands on the A$\to$B axis in SBERT space against the requested mix weight, with the diagonal marking proportional tracking. \doctolora and \texttt{ICAE} sit near a source paper at the endpoints and blend through the middle, while in-context prompting stays near the midpoint across the whole interior. \emph{Bottom}, verbatim copy rate. \texttt{ICAE} sits above the other two everywhere, and \doctolora and in-context prompting overlap. Paraphrase moves every curve by less than the gap between methods.}
\label{fig:psens-edge}
\end{figure}

We paraphrased the four-to-six sentence abstract prompt from \secref{fusion} three times. The wordings are listed as follows:
\begin{enumerate}[leftmargin=1.8em,itemsep=2pt,parsep=0pt,topsep=4pt]
\footnotesize
\item Write a detailed abstract (four to six sentences) describing this research topic: its problem, methods, and findings.
\item In four to six sentences, write a detailed abstract for this research topic, covering the problem it addresses, the methods it uses, and its findings.
\item Produce a detailed abstract of this research topic -- its problem, its methods and its findings -- in four to six sentences.
\item Describe this research topic as a detailed abstract of four to six sentences, stating the problem addressed, the methods used and the results obtained.
\end{enumerate}
For each wording, we decoded 20 near pairs and 20 far pairs at 13 mix weights for each method (\doctolora, in-context prompting, \texttt{ICAE}), totaling 6240 decodes. The in-context baseline used the same two-document template from Appendix~\ref{app:edge-protocol}, changing only the instruction.

Decoded abstracts closely tracked the requested mix for every wording (see \figref{psens-edge}). 
For \doctolora, slope of decoded position versus mix weight ranged from 1.11 to 1.20.
For \texttt{ICAE}, 1.19 to 1.24. In-context prompting ranged from 0.29 to 0.41. No wording changed the slope by more than 0.04.
Copy rate stayed consistent. \texttt{ICAE} copied the most, averaging $.72$ to $.75$, and never matched the other methods. 
\doctolora and in-context prompting were both near $.55$.

Changing the mix point rewrote abstracts much more than changing the prompt. Abstracts at the same point under different wordings had SBERT cosine similarity of 0.886 for \doctolora, 0.924 for in-context prompting, and 0.923 for \texttt{ICAE}. Abstracts at different points under one prompt had much lower cosine (0.146 for \doctolora, 0.256 for in-context prompting, 0.125 for \texttt{ICAE}).

\section{Benchmark performance of the baseline encoders with $g_\theta$}
\label{app:symmetric}

We used the same type of transformation (the same $g_\theta$ transform) in each baseline embedding space and repeated all $14$ benchmarks, following the same training steps as for \doctolora. 
However, the number of parameters in the transform varies between models: the \doctolora transform has $262{,}692$ parameters, while the transforms for the baseline encoders are larger---$590{,}593$ for text embeddings with $768$ dimensions, and $1{,}049{,}601$ for those with $1{,}024$ dimensions.

\begin{figure}[htbp]
\centering
\includegraphics[width=\textwidth]{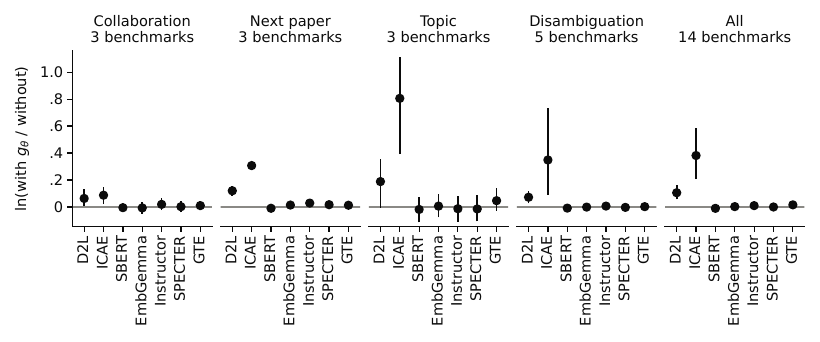}
\caption{Gain from the shared transform $g_\theta$ in every space on the $14$ benchmarks of \tabref{similarity}. For each benchmark and space we take the natural logarithm of the score with $g_\theta$ divided by the score without it. Each circle averages this log ratio over the benchmarks of a panel. The first four panels are collaboration prediction, next-paper prediction and topic classification on Physics, Economics and Psychology, and author-name disambiguation (B\textsuperscript{3} F1) on the five S2AND datasets. The last panel averages all $14$. Whiskers are $95\%$ intervals of a two-level bootstrap that resamples the panel's benchmarks and redraws each score with and without $g_\theta$ from its own bootstrap distribution. EmbGemma is \texttt{EmbeddingGemma}, and SPECTER is \texttt{SPECTER2} on the field benchmarks and \texttt{SPECTER} on S2AND, each with its own transform.}
\label{fig:symmetric-gain}
\end{figure}

\doctolora gains $+.061$ on average, improving on $11$ of the $14$ benchmarks (\figref{symmetric-gain} and \tabref{symmetric-adapter}). 
In contrast, baseline text encoders gain at most $+.009$ on average, and SBERT, the strongest baseline, loses $.007$ and degrades on $10$ of the $14$ benchmarks. 
In summary, for text encoders trained for similarity search, post adaptation via $g_\theta$ offers no recoverable structure and disrupts existing geometry. In \doctolora, it recovers latent citation-aligned geometry unexposed to cosine similarity.

\begin{table}[htbp]
\centering
\small
\caption{Change in benchmark score when the same transform $g_\theta$ is trained in each embedding space. Each row summarizes one space over the $14$ benchmarks of \tabref{similarity}. \emph{mean $\Delta$}: score with $g_\theta$ minus score without, averaged over the benchmarks. \emph{range}: smallest and largest change. \emph{better} and \emph{worse}: number of benchmarks with a significant gain or loss, i.e., a $95\%$ bootstrap interval of the change entirely above or below zero. Each bootstrap sample scores the space with and without $g_\theta$ on the same resampled evaluation units. This pairing yields a narrower interval for the change than the separate intervals of \tabref{similarity}.}
\label{tab:symmetric-adapter}
\begin{tabular}{lrrrr}
\toprule
space & mean $\Delta$ & range & better & worse \\
\midrule
\texttt{ICAE} & +.178 & [-.004, +.549] & 13/14 & 1/14 \\
\doctolora (Qwen3-4B) & +.061 & [-.009, +.133] & 11/14 & 1/14 \\
\texttt{Instructor} & +.009 & [-.020, +.040] & 7/14 & 2/14 \\
\texttt{GTE} & +.007 & [-.022, +.029] & 8/14 & 2/14 \\
\texttt{EmbeddingGemma} & +.001 & [-.044, +.025] & 6/14 & 3/14 \\
\texttt{SPECTER2} / \texttt{SPECTER} & +.001 & [-.034, +.023] & 5/14 & 4/14 \\
\texttt{SBERT} & -.007 & [-.017, +.005] & 1/14 & 10/14 \\
\bottomrule
\end{tabular}

\end{table}

\begin{table}[htbp]
\centering
\small
\caption{Scores of every space without $g_\theta$, averaged over the benchmarks of each task (three fields for collaboration, next-paper and topic, five S2AND datasets for disambiguation). Per-benchmark scores of \doctolora and the text encoders, with bootstrap standard deviations, are in \tabref{similarity}.}
\label{tab:symmetric-raw}
\begin{tabular}{lrrrr}
\toprule
space & Collab.\ (AUC) & Next paper (AUC) & Topic (F1) & Disamb.\ (B\textsuperscript{3} F1) \\
\midrule
\doctolora (Qwen3-4B) & .663 & .823 & .358 & .760 \\
\texttt{ICAE} (Mistral-7B) & .660 & .677 & .227 & .617 \\
\texttt{SBERT} & .726 & .941 & .443 & .827 \\
\texttt{EmbeddingGemma} & .712 & .919 & .433 & .813 \\
\texttt{Instructor} & .708 & .887 & .425 & .800 \\
\texttt{SPECTER2} / \texttt{SPECTER} & .708 & .915 & .417 & .784 \\
\texttt{GTE} & .711 & .925 & .434 & .816 \\
\bottomrule
\end{tabular}

\end{table}

\texttt{ICAE} is a method that reduces the amount of text fed into the context by compressing the input document into a fixed number of tokens called ``memory slots.''
It was not originally trained for representation learning, and our benchmarks confirm that it is the space with the lowest search performance (\tabref{symmetric-raw}).
However, as with \doctolora, the space is simply not structured to be used directly with cosine similarity.
The information useful for search is still contained within the space.
In fact, applying post-adaptation, as with \doctolora, improves performance by an average of $+.178$ and results in performance on par with \doctolora.
Across the $14$ benchmarks, the pairwise difference between the two adapted spaces averages $-.007$, with the maximum difference in either direction being $.056$ (Physics collaboration prediction, \texttt{ICAE} ahead).
Therefore, once both compression spaces have $g_\theta$, the two spaces exhibit virtually equivalent performance in terms of retrieval scores.
What distinguishes the two is their decoding behavior: \texttt{ICAE} reproduces a higher proportion of the source text (\figref{cluster-labels}f), and its cluster labels underperform \doctolora with a score of $.60$ in the blind evaluation (\secref{cluster-labeling}).

\section{Full result tables}
\label{app:tables}

This appendix collects the tables that the preceding appendices read against.
Every table is generated by the workflow from the same caches as the figures in the main text.

\tabref{encoder-matrix} repeats the field tasks for the three base models of Appendix~\ref{app:benchmarks}.
\tabref{collab-per-window} splits the collaboration AUC of \tabref{similarity} into its anchor windows.
\tabref{radial-vec2text} is the \texttt{vec2text} magnitude sweep of Appendix~\ref{app:length-dial}.
\tabref{hierarchy-labels} closes the appendix with the native output of every embedding method for every node of the PACS slice scored in Appendix~\ref{app:baseline-labels}.

\begin{table}[htbp]
\centering
\small
\caption{Per-encoder field-task matrix. Each cell reports the two \doctolora variants as raw\,/\,+$g_\theta$ (the frozen embedding and the single general citation transform), with metrics, evaluation pools, and bootstrap protocol as in \tabref{similarity} (standard deviations omitted). Text-baseline scores are encoder-independent and given in \tabref{similarity}.}
\label{tab:encoder-matrix}
\begin{tabular}{llccc}
\toprule
Encoder & Field & Next-paper (AUC) & Topic (F1) & Collaboration (AUC) \\
\midrule
\multirow{3}{*}{Gemma-2-2b}
 & Economics & .792\,/\,.888 & .217\,/\,.317 & .592\,/\,.617 \\
 & Psychology & .786\,/\,.910 & .180\,/\,.319 & .593\,/\,.673 \\
 & Physics & .864\,/\,.946 & .569\,/\,.582 & .827\,/\,.823 \\
\midrule
\multirow{3}{*}{Qwen3-4B}
 & Economics & .813\,/\,.910 & .242\,/\,.374 & .596\,/\,.629 \\
 & Psychology & .799\,/\,.923 & .217\,/\,.340 & .602\,/\,.687 \\
 & Physics & .880\,/\,.958 & .590\,/\,.590 & .792\,/\,.795 \\
\midrule
\multirow{3}{*}{Mistral-7B}
 & Economics & .825\,/\,.913 & .234\,/\,.368 & .599\,/\,.639 \\
 & Psychology & .815\,/\,.926 & .217\,/\,.348 & .601\,/\,.681 \\
 & Physics & .876\,/\,.965 & .591\,/\,.595 & .820\,/\,.781 \\
\bottomrule
\end{tabular}

\end{table}

\begin{table}[htbp]
\centering
\small
\setlength{\tabcolsep}{4pt}
\caption{AUC-ROC of collaboration prediction for each anchor year. At an anchor year, each author is represented by the mean embedding of their papers from the previous four years. Candidate pairs are authors two steps apart in the coauthorship network who have not published together. A candidate pair is positive if the two authors coauthor a paper within the three years after the anchor year. Each column scores the candidate pairs of one anchor year. The \emph{pooled} column scores the candidate pairs of all anchor years together and is the number reported in \tabref{similarity}. \doctolora is the Qwen3-4B embedding with and without $g_\theta$ (Appendix~\ref{app:benchmarks}).}
\label{tab:collab-per-window}
\resizebox{\textwidth}{!}{\begin{tabular}{lrrrrrrrrrrr}
\toprule
method & \multicolumn{3}{c}{Physics} & \multicolumn{4}{c}{Economics} & \multicolumn{4}{c}{Psychology} \\
\cmidrule(lr){2-4}\cmidrule(lr){5-8}\cmidrule(lr){9-12}
 & 2000 & 2008 & pooled & 2008 & 2012 & 2016 & pooled & 2008 & 2012 & 2016 & pooled \\
\midrule
\doctolora (raw) & .786 & .848 & .792 & .646 & .556 & .579 & .596 & .581 & .610 & .627 & .602 \\
\quad $+g_\theta$ & .784 & .849 & .795 & .680 & .594 & .603 & .628 & .672 & .721 & .678 & .688 \\
\texttt{SPECTER2} & .788 & .857 & .826 & .672 & .593 & .621 & .632 & .651 & .679 & .676 & .665 \\
\texttt{SBERT} & .826 & .887 & .832 & .696 & .626 & .637 & .655 & .691 & .718 & .667 & .691 \\
\texttt{Instructor} & .831 & .843 & .854 & .676 & .577 & .597 & .620 & .645 & .657 & .648 & .649 \\
\texttt{EmbeddingGemma} & .798 & .908 & .829 & .684 & .596 & .612 & .633 & .663 & .688 & .652 & .667 \\
\texttt{GTE} & .814 & .825 & .838 & .670 & .593 & .610 & .627 & .665 & .682 & .659 & .668 \\
\bottomrule
\end{tabular}
}
\end{table}

\begin{table}[htbp]
\centering
\footnotesize
\setlength{\tabcolsep}{3pt}
\caption{\texttt{vec2text} magnitude sweep on the same five seed documents. \texttt{vec2text} is an inverter and takes no prompt, so we read back its reconstruction directly. The reconstruction stays on topic at the same surface specificity for larger $\alpha$ and degrades into off-topic fragments at small $\alpha$, never abstracting toward the broader field. Entries truncated to leading terms.}
\label{tab:radial-vec2text}
\rowcolors{2}{gray!12}{white}
\begin{tabularx}{\linewidth}{@{}c *{5}{>{\raggedright\arraybackslash}X}@{}}
\toprule
$\alpha$ & \textbf{Physics} & \textbf{Biochemistry} & \textbf{Fruit} & \textbf{Person} & \textbf{Patent (LED)} \\
\midrule
$1.15$ & is the fractional quantum effect of accumulating electrons in certain atoms.\,\dots & (abbreviated CY), commonly ATP, is an enzyme cycle for the production\,\dots & a banana is an edible plantation fruit, traditionally specializing in long,\,\dots & Nikola Tesla, an American engineer from Serbia, was an 1860s American\,\dots & comprised an ion-doped light-emitting semiconductor forming a nitride structure, nitrid \\
$0.9$ & quantum fraction is a function of the Hall effect in physics:\,\dots & enzyme cycle is the metabolic use of ATP for generating cryptic\,\dots & fruit is a elongated banana. This is due to the variety\,\dots & Nikola Tesla was a Serbian engineer and American engineer, known for\,\dots & structure comprised a light-emitting compound of a double-type nitride semiconductor (e.g. \\
$0.7$ & quantum mechanics is the fraction of the Hall effect which is\,\dots & acids used for the production of glycogen. The Krebs cycle is\,\dots & plant species is a banana. The name derives from a linguistic\,\dots & engineering genius Nikola Tesla is a Serbian engineer who contributed to\,\dots & structure of a double-layered nickel semiconductor (indicated in the introduction of\,\dots \\
$0.55$ & equation for the physics of fractional heating is a contribution to\,\dots & the CYCY process of carbohydrates is a way of generating ATP,\,\dots & botanical designation of a plant. A banana is a combination of\,\dots & American mechanical engineer. Nikola Tesla is credited for the production of\,\dots & layer of a g-type nickel plating (indicated in the following table:\,\dots \\
$0.42$ & equation for quantum mechanics is a result of the publication of\,\dots & the metabolic cycle of enzymes is a contribution to the aforementioned\,\dots & botanical name for a banana. This is a variant of the\,\dots & American electrical engineer. Nikola Tesla is credited for the contribution of\,\dots & a GLTI semiconductor pairing of the following structures: (a) (initiated) (a)\,\dots \\
$0.3$ & a number of awards in the field of physics & biology is a key component of the Crypt of the Chemical\,\dots & biology of plants is a classic in the discipline of banana\,\dots & American electrical engineer who was a major contributor to the creation\,\dots & a layered semiconductor for the neo-Neo-Geo-Neo-Neo-Dai \\
$0.2$ & a prestigious award in the field of physics for the 1860\,\dots & biology is a classic in the vein of the MIT CYPATHE,\,\dots & classic in Asian agriculture. The title of a chapter is a\,\dots & American mechanical engineer. Tesla contributed to a number of awards including\,\dots & a DP-type semiconductor engineering is \\
$0.12$ & a prestigious journal in the area of physics entitled & a.k.a. The Chemical Cycle of the As - a -does it\,\dots & classic in the history of the Venus Project & a history of Tesla engineering. Contributor to the 97th edition of\,\dots & a fusion of the following disciplines \\
$0.06$ & classic in the history of physics. 109. & a.k.a. 179. 2004: The Academy of Sciences published a review of\,\dots & classic in the history of engineering. William Rice Publishers & 1925. The publication of a book on the history of electricity\,\dots & a.k.a. DP Engineering \\
\bottomrule
\end{tabularx}
\end{table}

\begin{footnotesize}
\setlength{\tabcolsep}{3pt}
\renewcommand{\arraystretch}{1.2}
\rowcolors{3}{gray!12}{white}
\begin{longtable}{@{}>{\raggedright\arraybackslash}p{1.85cm} >{\raggedright\arraybackslash}p{2.55cm} >{\raggedright\arraybackslash}p{2.6cm} >{\raggedright\arraybackslash}p{2.55cm} >{\raggedright\arraybackslash}p{2.55cm}@{}}
\caption{
Cluster labels for every node of a four-branch PACS slice. Each cell is the method's native output, exactly the string scored in \tabref{label-eval}, truncated at a word boundary for the page but scored in full. \doctolora and \texttt{ICAE} answer the field prompt; \texttt{KeyLLM} returns a keyword list and \texttt{vec2text} a reconstruction, which is why only the first two columns read as field names. The first column gives the PACS code and a short name, indented by tree depth, with field rows in bold.}
\label{tab:hierarchy-labels} \\
\toprule
\rowcolor{white}
& \multicolumn{4}{c}{\textbf{Native output (as scored)}} \\
\cmidrule(lr){2-5}
\rowcolor{white}
\textbf{PACS node} & \textbf{\doctolora} & \textbf{\texttt{KeyLLM}} & \textbf{\texttt{ICAE}} & \textbf{\texttt{vec2text}} \\
\midrule
\endfirsthead
\toprule
\rowcolor{white}
& \multicolumn{4}{c}{\textbf{Native output (as scored)}} \\
\cmidrule(lr){2-5}
\rowcolor{white}
\textbf{PACS node} & \textbf{\doctolora} & \textbf{\texttt{KeyLLM}} & \textbf{\texttt{ICAE}} & \textbf{\texttt{vec2text}} \\
\midrule
\endhead
\bottomrule
\endlastfoot
\textbf{0~~General} & \textbf{Quantum field theory} & partial decoherence, thermalization, time-domain\,\dots & Quantum many-body systems & coupled-dissonance correlations in one-way systems. The\,\dots \\
\hspace{0.8em}03~~Quantum mechanics, QFT & \textbf{Quantum optics} & quantum state transfer, entanglement, four qubits\,\dots & Quantum optics & coupled entanglement of one-state quantums, where the\,\dots \\
\hspace{1.8em}03.65~~Quantum mechanics & \textbf{Quantum mechanics} & quantum decoherence, single qubit, random matrix theory\,\dots & Quantum mechanics & as an empirical measure of the interpolation of one-state\,\dots \\
\hspace{1.8em}03.67~~Quantum information & \textbf{Quantum information} & Measurement-based quantum computing, AKLT state, spin-1\,\dots & Quantum information & entanglement of single-state quantum computations, so\,\dots \\
\hspace{0.8em}05~~Statistical physics & \textbf{Statistical physics} & Disturbance spreading, Incommensurate systems\,\dots & Statistical mechanics & diffuse-phase dynamics in univariate quantization. This\,\dots \\
\hspace{1.8em}05.40~~Fluctuations, noise & \textbf{Statistical physics} & bistable systems, colored noise, stochastic relaxation\,\dots & Statistical physics & diffusion-display coefficients in one-variable\,\dots \\
\hspace{1.8em}05.45~~Nonlinear dynamics, chaos & \textbf{Nonlinear dynamics} & separatrix map, resonant dynamics, global chaos onset\,\dots & Nonlinear dynamics & coupled chaotic quantization in classical dynamics. This\,\dots \\
\addlinespace[2pt]
\textbf{2~~Classical physics} & \textbf{Nonlinear optics} & Nonlinear light pulse propagation, Polarized light\,\dots & Quantum optics & multi-mode scattering of neutrons via classical optical\,\dots \\
\hspace{0.8em}42~~Optics & \textbf{Quantum optics} & Photon antibunching, Photonic molecule, Coupled cavities\,\dots & Quantum optics & coupled photon-diametric scattering of neutrons in\,\dots \\
\hspace{1.8em}42.50~~Quantum optics & \textbf{Quantum optics} & mesoscopic states, cavity-field state, atom-field\,\dots & Quantum optics & two-level physics involving scattering of coherent atoms\,\dots \\
\hspace{1.8em}42.65~~Nonlinear optics & \textbf{Nonlinear optics} & solitons, spontaneous symmetry breaking, photonic\,\dots & Nonlinear optics & nonsimultaneous low-wave propagation in coherent optical\,\dots \\
\hspace{0.8em}47~~Fluid dynamics & \textbf{Fluid dynamics} & Two-dimensional turbulence, freely decaying turbulence\,\dots & Fluid dynamics & fluid-displacement dynamics in experimentally correlated\,\dots \\
\hspace{1.8em}47.20~~Flow instabilities & \textbf{Fluid dynamics} & Bifurcation diversity, Thermocapillary liquid layers\,\dots & Fluid dynamics & in-convection instability in linear-wavelength fluid\,\dots \\
\hspace{1.8em}47.27~~Turbulent flows & \textbf{Fluid dynamics} & Mean-field approximation, turbulence theory\,\dots & Turbulence theory & in turbulence-displacement dynamics, the initial\,\dots \\
\addlinespace[2pt]
\textbf{3~~Condensed matter} & \textbf{Condensed matter physics} & Quasi-one-dimensional electron systems, t-J model\,\dots & Condensed matter physics & in-distance interactions between electron-bonding\,\dots \\
\hspace{0.8em}71~~Electronic structure & \textbf{Condensed matter physics} & semiclassical action, dynamical mean-field theory, local\,\dots & Condensed matter physics & phase-dimersion configuration of classical electron-fluid\,\dots \\
\hspace{1.8em}71.10~~Many-electron systems & \textbf{Condensed matter physics} & Mott physics, Mott transition, spin fluctuations\,\dots & Condensed Matter Physics & one-dimensional quasi-dotonal model where the Hubbard\,\dots \\
\hspace{1.8em}71.20~~Bulk band structure & \textbf{Condensed matter physics} & Electronic structure, Pyrochlore metals, Cd2Os2O7\,\dots & Materials science & (dBa)-structure calculations of the respective metallic\,\dots \\
\hspace{0.8em}75~~Magnetic materials & \textbf{Condensed matter physics} & Quantum Monte Carlo, sign problem, itinerant\,\dots & Condensed Matter Physics & inferromagnetic spin-diabetic scattering interactions\,\dots \\
\hspace{1.8em}75.10~~Magnetic ordering & \textbf{Condensed matter physics} & square-lattice Heisenberg antiferromagnet, S=1/2\,\dots & Condensed Matter Physics & in anastomosing-phase-spatial-inverse quantum state, with\,\dots \\
\hspace{1.8em}75.30~~Ordered magnets & \textbf{Condensed matter physics} & metal-insulator transition, La1-xCaxMnO3, manganites\,\dots & Magnetism & inferomorphic magnetic spin-phases at 0–0,0–0,0–0,0–0,0–0 \\
\addlinespace[2pt]
\textbf{6~~Elementary particles} & \textbf{Particle physics} & Left-right symmetry, LHC, SU(2)\_L x SU(2)\_R x U(1)\_B-L\,\dots & High Energy Physics & scattering of resonant singularity quarks on the\,\dots \\
\hspace{0.8em}11~~Fields and particles & \textbf{Quantum field theory} & light-front Hamiltonian, spontaneous chiral symmetry\,\dots & Quantum Field Theory & with the symmetries of fermion quantum diffusion. This\,\dots \\
\hspace{1.8em}11.10~~Field theory & \textbf{Quantum field theory} & sine-Gordon field theory, massive, (1+1)-dimensional\,\dots & Quantum field theory & symmetries of the classical quantization-dynamic theory\,\dots \\
\hspace{1.8em}11.15~~Gauge field theories & \textbf{Quantum field theory} & SU(2) gauge theory, adjoint Dirac flavor, infrared\,\dots & Quantum field theory & scattering of d-quantity quantums. This d-quantity\,\dots \\
\hspace{0.8em}12~~Particle systematics & \textbf{Particle physics} & Left-Right symmetry, LHC, SU(2)\_L x SU(2)\_R x U(1)\_B-L\,\dots & High Energy Physics & scattering of low-mass quark quantities. This approach\,\dots \\
\hspace{1.8em}12.38~~Quantum chromodynamics & \textbf{Quantum chromodynamics} & Nucleon strange quark content, Lattice QCD, Two-flavor\,\dots & Nuclear physics & with d-quantity scattering of d-quantity laton particles\,\dots \\
\hspace{1.8em}12.60~~Beyond standard model & \textbf{Particle physics} & SO(10), supersymmetry breaking, gauge mediation, grand\,\dots & Particle physics & high scattering-symmetry sbN, the experimental model will\,\dots \\

\end{longtable}
\end{footnotesize}

\end{document}